\documentclass{siamart190516}%
\usepackage{lineno,hyperref}
\usepackage{lipsum}
\usepackage{amsfonts, amssymb}
\usepackage{graphicx}
\usepackage{epstopdf}
\usepackage{algorithmic}
\newtheorem{thm}{Theorem}
\newtheorem{asm}{Assumption}
\newtheorem{Rem}{Remark}
\newcommand{\dd}[2]{\frac{\d #1}{\d #2}}
\renewcommand{\d}{\ensuremath{\,\mathrm{d}}}
\renewcommand{\vec}[1]{\ensuremath{\boldsymbol{#1}}}
\newcommand{\abs}[1]{\left| #1 \right|}
\newcommand{\norm}[1]{\left\|{#1}\right\|}
\DeclareMathOperator*{\supp}{supp}

\begin{document}
\nolinenumbers

\title{A Phase Shift Deep Neural Network for High Frequency Approximation and Wave Problems \thanks{to be submitted to SISC.}}
\author{Wei Cai\thanks{Department of Mathematics, Southern Methodist University, Dallas, TX 75275, USA.}\and{Xiaoguang Li\thanks{MOE-LCSM, School of Mathematics and Statistics, Hunan Normal University, Changsha, Hunan410081, P. R. China}}\and{LiZuo Liu\thanks{Department of Mathematics, Southern Methodist University, Dallas, TX 75275, USA.}}}
\date{December 4, 2019}
\maketitle
%
%\begin{frontmatter}
%	\title{A Phase Shift Deep Neural Network for High Frequency Approximation and Wave Problems}
%	
%	%% Group authors per affiliation:
%\address[add1]{Department of Mathematics, Southern Methodist University, Dallas, TX 75275, USA.}
%	\address[add2]{LCSM, Ministry of Education, School of Mathematics and Statistics, Hunan Normal University, Changsha, Hunan 410081, P. R. China}
%\author[add1]{Wei Cai}
%%\corref{mycorrespondingauthor}
%%\cortext[mycorrespondingauthor]{Corresponding author}
%%\ead{cai@smu.edu}}
%	\author[add2,add1]{Xiaoguang Li}
%\author[add1]{Lizuo Liu}

	\begin{abstract}
		In this paper, we propose a phase shift deep neural network (PhaseDNN), which provides a uniform wideband convergence in approximating high frequency functions and solutions of wave equations. The PhaseDNN makes use of the fact that common DNNs often achieve convergence in the low frequency range first, and a series of moderately-sized DNNs are constructed and trained for selected high frequency ranges.  With the help of phase shifts in the frequency domain, each of the DNNs will be trained to approximate the function's higher frequency content over a specific range at the the speed of convergence as in the low frequency range. As a result, the proposed PhaseDNN is able to convert high frequency learning to low frequency one, allowing a uniform learning to wideband functions. The PhaseDNN will then be applied to find the solution of high frequency wave equations in inhomogeneous media through both differential and integral equation formulations with least square residual loss functions. Numerical results have demonstrated the capability of the PhaseDNN in learning high frequency functions and oscillatory solutions of interior and exterior Helmholtz equations.
	\end{abstract}
	

%\end{frontmatter}

\section{Introduction}

Deep neural networks (DNNs) have shown greater potential in approximating high
dimensional functions, compared with traditional approximations based on
Lagrangian interpolation or spectral methods. Recently, it has been found
\cite{xu18, xu19, xu2019theory} that some common NNs, including fully connected and
convolution neural network (CNN) with tanh and ReLU activation functions,
demonstrate a frequency dependent convergence behavior. Namely, the DNNs
during the training are able to approximate the low frequency components of
the targeted functions first before higher frequency components. This phenomena is defined as
the F-Principle of DNNs \cite{xu18}. The stalling of DNN convergence
in the later stage of training could be mostly related to learning the high
frequency components of the data. The F-principle behavior of DNNs is the
opposite to that of the traditional multigrid method (MGM) \cite{bandt77} in
approximating the solutions of PDEs where the convergence occurs first in the
higher frequency end of the spectrum, as a result of the smoothing operator employed
in the MGM. The MGM takes advantage of this fast high frequency error
reduction in the smoothing iteration cycles and restricts the original solution
on a fine grid to a coarser grid, then continuing the smoothing iteration on
the coarse grid to reduce the higher end frequency spectrum in the context of
the coarse grid. This downward restriction can be continued until errors over
all frequency are reduced by a small number of iterations on each level of the
coarse grids.

%By reformulating a differential equation as a optimization problem and approximating the solution by a DNN parameterized by $\mathbf{\theta}$
%one can solve differential equation by DNNs\cite{RitzDNN},\cite{raissi2019physics}. The Physics-informed neural networks(PINN) proposed in \cite{raissi2019physics} has the advantage that it is quite flexible and easy to implement. It has been successfully applied to elliptic PDE. However, due to the F-principle, there is some difficulties in using PINN to solve wave equations with high frequency.

There are many scientific computing problems which involve high frequency solutions,
such as high frequency wave equations in inhomogeneous media, arising from electromagnetic wave propagation
in turbid media,  seismic waves, and geophysical problems. Finding efficient solutions, especially
in random environments,  poses great
computational challenges due to the highly oscillatory natures of the
solutions as well as the variance of random media properties. To model the
high frequency waves in a deterministic media, high order methods such as
spectral methods for the differential equations or wideband fast multipole
methods for the integral equations are often used. Sparse grid methods have been
used to reduce the cost of simulation over the high dimensional random spaces,
however, they still suffer the curse of dimensionality.
In this paper, we will  develop DNNs based numerical methods to handle high frequency
functions and solutions of high frequency wave equation in inhomogeneous
media, whose solution could be oscillatory as well as high dimensional due
to the random coefficients. Our goal is to develop new
classes of DNNs with capability of representing highly oscillatory solutions
in the physical spatial variables, and  at the same time addressing the
challenges of representing high random space dimensions, eventually.

To improve the capability of usual DNNs for learning highly oscillatory
functions in the physical spatial variables, we propose a phase shift DNN
with wideband learning capabilities in error reductions in the approximation
for all frequencies of the targeted function by taking advantage of the faster
convergence in the low frequencies of the DNN during its training.
%We approximate the target function with a series of functions with disjoint
%frequency spectrum through a partition of the unit (POU) in the $k$-space.
To learn a function of specific frequency range, we employ a phase
shift in the $k$-space to translate its frequency to the range $|k|<K_{0}$,
then the phase shifted function with a low frequency content can be
learned by common DNNs with a small number of training epoches. The resulting series of DNNs
with phase shifts will make a phase shift deep neural network (PhaseDNN).

To achieve uniform wideband approximation of a general function, we can implement the
PhaseDNN in a parallel manner where original data is decomposed into data of
specific frequency range, which after a proper phase shift, is learned quickly. This approach can
be implemented in a parallel manner, however, frequency extraction of
the original training data have to be done using convolutions with a frequency
selection kernel numerically, which could become very expensive or not accurate for
scattered training data. Alternatively, we can implement the PhaseDNN in
a non-parallel manner where data from all range of frequencies are learned
together with phase shifts included in the makeup of the PhaseDNN, resulting in a
coupled PhaseDNN. Although, the coupled PhaseDNN lacks parallelism, it avoids the costly convolution used in the parallel
PhaseDNN to extract the frequency component from the original training data.
This feature will be shown to be important when higher dimensional data are
involved in the training. Thanks to this property, the coupled PhaseDNN will be
used to solve high frequency wave problems where we seek solutions in
a space of PhaseDNNs by minimizing the residuals of the differential equation in
a least square approach.

The rest of the paper will be organized as follows. In section 2, we will
review the fast low frequency convergence property of neural network and  present the parallel version phase shift
deep neural network - PhaseDNN. Based on the properties of the PhaseDNN, a
coupled PhaseDNN is introduced in Section 3 to reduce the cost of learning in
training the DNN for approximations. Then, the coupled PhaseDNN is used to find the solutions of
wave problems in inhomogeneous media using either differential equation or integral equation formulations.
Section 4 contains
various numerical results of the PhaseDNN for approximations and solutions of
wave problems. A conclusion and discussions
will be given in Section 5. Appendix will include an analysis of the equivalence between the parallel
PhaseDNN and coupled PhaseDNN, providing the theoretical basis for the accurate results of coupled PhaseDNN.

\section{A parallel phase shift DNN (PhaseDNN) for high frequency approximation}

A deep neural network (DNN) is a sequential alternative composition of linear
functions and nonlinear activation functions. Given $m,n\geq1$, let
$\Theta(\vec{x}):\mathbb{R}^{n}\rightarrow\mathbb{R}^{m}$ is a linear function
with the form $\Theta(\vec{x})=\vec{W}\vec{x}+\vec{b}$, where $\vec{W}%
=(w_{ij})\in\mathbb{R}^{m\times n}$, $\vec{b}\in\mathbb{R}^{m}$ are called weights and biases, respectively. The nonlinear
activation function $\sigma(u):\mathbb{R}\rightarrow\mathbb{R}$. By applying $\sigma(u)$
componentwisely, we can extend the activation function to
$\sigma(u):\mathbb{R}^{n}\rightarrow\mathbb{R}^{n}$. A DNN with
$L+1$ layers can be expressed in a compact form as
\begin{equation}
\begin{aligned} T(\vec{x}) &= T^L(\vec{x}),\\ T^l(\vec{x}) & = [\Theta^l\circ\sigma](T^{l-1}(\vec{x})), \quad l=1,2,\dots L, \end{aligned} \label{eq:dnniter}%
\end{equation}
with $T^{0}(\vec{x})=\Theta^{0}(\vec{x})$, or equivalently,
it explicitly:
\begin{equation}
T(\vec{x})=\Theta^{L}\circ\sigma\circ\Theta^{L-1}\circ\sigma\cdots\circ
\Theta^{1}\circ\sigma\circ\Theta^{0}(\vec{x}). \label{eq:dnn}%
\end{equation}
Here, $\Theta^{l}(\vec{x})=\vec{W}^{l}\vec{x}+\vec{b}^{l}:\mathbb{R}^{n^{l}%
}\rightarrow\mathbb{R}^{n^{l+1}}$ are linear functions. This DNN is also said
to have $L$ hidden layers and its $l$-th layer has $n^{l}$ neurons.

In approximating a function $f(x)$ by a DNN through training, we  minimize the least square loss function
\begin{equation}
L(\vec{W}^{0},\vec{b}^{1},\vec{W}^{1},\vec{b}^{1},\dots,\vec{W}^{L},\vec
{b}^{L})=\left\Vert {f(\vec{x})-T(\vec{x})}\right\Vert _{2}^{2}=\int_{-\infty
}^{+\infty}\left\vert f(\vec{x})-T(\vec{x})\right\vert ^{2}\d x. \label{eq:loss}%
\end{equation}
For simplicity, we denote all the parameters in DNN by a parameter vector $\theta$, i.e.
\[
\theta=(\vec{W}_{11}^{0},\dots,\vec{W}_{n^{0}n^{1}}^{0},\vec{b}_{1}^0\dots \vec{b}^{0}_{n^{1}}%
,\vec{W}_{11}^{1},\dots,\vec{W}_{n^{1} n^{2}}^{1},\vec{b}_{1}^{1}\dots
\vec{b}^{1}_{n^{2}}\dots)\in\mathbb{R}^{p}.
\]
Here, $p=(n^{0}+1)\times n^1+(n^{1}+1)\times n^{2} + (n^{2}+1)\times n^{3}+\dots(n^{L}+1)$
is the total number of the parameters. Numerically, with $N$ training data $\{x_1,x_2,\dots, x_N\}$, the numerical loss function is defined as
\begin{equation}\label{eq:lossN}
  L_N(\theta) = \sum_{i=1}^{N}\abs{f(x_i)-T(x_i,\theta)}^2.
\end{equation}

We can study the loss function in the frequency space and first,  define the Fourier transform and its inverse of a function $f(\vec{x})$ by
\begin{equation}
\label{eq:fourier}\mathcal{F}[f](k)=\frac{1}{\sqrt{2\pi}}\int_{-\infty
}^{+\infty}f(x)e^{-ikx}\d x,\quad\quad\mathcal{F}^{-1}[\hat{f}](x) =
\frac{1}{\sqrt{2\pi}}\int_{-\infty}^{+\infty}\hat{f}(k)e^{ikx}\d k.
\end{equation}

Assuming the Fourier transform of $f(x)$ and $T(x,\theta)$ exist, by Paseval's equality, we have
\begin{equation}
L(\theta)=\int_{-\infty}^{+\infty}|f(x) - T(x)|^2\d x = \int_{-\infty}^{+\infty}|\hat{f}(x) - \hat{T}(x)|^2\d k. \label{eq:lossfourier}%
\end{equation}

The F-principle states the relative changing rate of $L(\theta)$ and $\hat{T}$ along the training trajectory for different frequency component. Specifically, for a constant $\eta>0$, let
\begin{equation}\label{eq:Linandout}
  L_\eta^-(\theta) = \int_{B_\eta}|\hat{f}(x) - \hat{T}(x)|^2\d k, \quad L_\eta^+(\theta) = \int_{B_\eta^c}|\hat{f}(x) - \hat{T}(x)|^2\d k,
\end{equation}
where $B_\eta = \{x|\norm{x}<\eta\}$ is the ball with radius $\eta$, $B_\eta^c$ is its complement. The following result is proven in \cite{xu2019theory}.

\begin{thm}\label{thm:FP}
Suppose the training process of a DNN in (\ref{eq:dnn}) is carried out by a gradient decent method, i.e.,
\[\dd{\theta}{t} = -\nabla_{\theta} L(\theta).\]
If on the training trajectory, the following assumptions hold:
\begin{enumerate}
  \item$f(x), T(x)\in H^{r}(\mathbb{R}^d)$ for $r\geq 1$,
  \item $\theta(t)\neq constant$ and $\sup_{t\geq 0}\abs{\theta(t)}\leqslant R$ for a constant $R>0$,
  \item $\inf_{t>0}|\nabla_{\theta}L(\theta)|>0$,
\end{enumerate}
then, in a fixed training time interval $t\in [0,T]$, there exists a constant $C>0$ such that
\begin{equation}\label{FPforL}
  \frac{|\d L^+_\eta(\theta)/\d t|}{|\d L(\theta)/\d t|} <C\eta^{-m}, \quad\mbox{and }\quad \frac{|\d L^-_\eta(\theta)/\d t|}{|\d L(\theta)/\d t|} >1 - C\eta^{-m} \quad \forall t\in [0,T]
\end{equation}
for all $1\leqslant m\leqslant 2r-1$. Moreover,
\begin{equation}\label{FPforT}
  \frac{\norm{\d \hat{T}/\d t}_{L^2(B^c_\eta)}}{\norm{\d \hat{T}/\d t}_{L^2(\mathbb{R}^d)}} <C\eta^{-m}, \quad\mbox{and }\quad \frac{\norm{\d \hat{T}/\d t}_{L^2(B_\eta)}}{\norm{\d \hat{T}/\d t}_{L^2(\mathbb{R}^d)}} >1 - C\eta^{-m}\quad \forall t\in [0,T]
\end{equation}
for all $1\leqslant m\leqslant r-1$.
\end{thm}
%
%%\begin{Rem}
%%The proof of this theorem should be separated into several parts. First we
%%should make sure the existence of $\mathcal{F}[T](k)$. Secondly, cite Xu
%%zhiqing's work as the proof of one-layer case. Thirdly, show the dominate
%%exponential estimation inductively.
%%\end{Rem}
%
%
%\section{A parallel phase shift DNN (PhaseDNN)}

%So when we train a DNN $T_{\star}(x)$ to approximate $f(x)$, there exists positive
%frequency number $K_{0}$ and an integer $n_{0}$ such that after $n_{0}$ steps
%of training, the Fourier transform of training function $\mathcal{F}[T_{\star
%}](\vec{k};\theta^{(n_{0})})$ should be a good approximation of $\mathcal{F}%
%[f](\vec{k})$ for $\left\vert k\right\vert <K_{0}$, where $\theta^{(n_{0})}$
%are parameters obtained after $n_{0}$ steps of training.

The F-principle and estimations states that when gradient decent method is applied to loss function, the low frequency part of loss function converges faster than the high frequency part. Therefore, as in \cite{phaseShift}, to speed up the learning of the higher frequency content of the target function $f(x)$, we can employ a phase shift technique to translate higher frequency spectrum $\hat{f}(k)$ to the frequency range of $[-K_{0},K_{0}]$ for a small frequency $K_{0}$. Such a shift in frequency is a simple phase factor multiplication on the training data in the physical space.

\subsection{Frequency selection kernel $\phi_{j}^{\vee}(x)$}
For a given frequency increment $\Delta k$, say, $\Delta k = 2K_0$, let us assume that for some integer $M>0$,
\[
\supp\hat{f}(k)\subset\lbrack-M\Delta k,M\Delta k].
\]
We first construct a mesh for the interval
$[-M\Delta k,M\Delta k]$ by
\begin{equation}
\omega_{j}= j\Delta k,j=-M,\cdots,M, \label{mesh}%
\end{equation}

Then, we introduce a POU (partition of unit) $\{\phi_{j}(k)\}_{j=-M}^M$ for the interval $[-M\Delta k,M\Delta k]$ associated with the mesh as%
\begin{equation}
1=
{\displaystyle\sum\limits_{j=-M}^{M}}
\phi_{j}(k),\text{ }k\in\lbrack-M\Delta k,M\Delta k]. \label{pou}%
\end{equation}
The simplest choice of $\phi_{j}(k)$ is
$
\phi_{j}(k)=\phi(\frac{k-\omega_{j}}{\Delta k}),
$
%\begin{equation}
%\label{eq:phij}\phi_{j}(k)=\phi(\frac{k-\omega_{j}}{\Delta k}),
%\end{equation}
and $\phi(k)$ is just the characteristic function of $[-\frac{1}{2},\frac{1}{2}],$ i.e., $\phi(k)=\chi_{\lbrack -\frac{1}{2},\frac{1}{2}]}(k).$
%\[
%\phi(k)=\chi_{\lbrack -\frac{1}{2},\frac{1}{2}]}(k).
%\]
The inverse Fourier transform $\mathcal{F}^{-1}$ of $\phi(k)$, indicated by $\vee,$ is $\phi^{\vee}(x) = \frac{1}{\sqrt{2\pi}}\frac{\sin \frac{x}{2}}{\frac{x}{2}}.$
%\begin{equation}
%\label{eq:invfphi}\phi^{\vee}(x) = \frac{1}{\sqrt{2\pi}}\frac{\sin \frac{x}{2}}{\frac{x}{2}}.
%\end{equation}

With the POU in (\ref{pou}), we can decompose the target
function $f(x)$ in the Fourier space as follows,%
\begin{equation}
\hat{f}(k)=%
%TCIMACRO{\dsum \limits_{j=-m}^{m}}%
%BeginExpansion
{\displaystyle\sum\limits_{j=-M}^{M}}
%EndExpansion
\phi_{j}(k)\hat{f}(k)\triangleq%
%TCIMACRO{\dsum \limits_{j=-m}^{m}}%
%BeginExpansion
{\displaystyle\sum\limits_{j=-M}^{M}}
%EndExpansion
\hat{f_{j}}(k), \label{k-decomp}%
\end{equation}
%where%
%\[
%\sup\hat{f_{j}}(k)\subset\lbrack\omega_{j-1,}\omega_{j+1}].
%\]
which will give a corresponding decomposition in $x$-space as%
\begin{equation}
f(x)=%
%TCIMACRO{\dsum \limits_{j=-m}^{m}}%
%BeginExpansion
{\displaystyle\sum\limits_{j=-M}^{M}}
%EndExpansion
f_{j}(x), \label{x-decomp}%
\end{equation}
where
\[
f_{j}(x)=\mathcal{F}^{-1}[\hat{f_{j}}](x).
\]

The decomposition (\ref{x-decomp}) involves $2M+1$ functions $f_{j}(x)$, whose
frequency spectrum is limited to $[\omega_{j}-\frac{\Delta k}{2}, \omega_{j}+\frac{\Delta k}{2}]$, therefore, a
simple phase shift could translate its spectrum to $[-\Delta k/2,\Delta k/2]$, and it
could be learned quickly by a relatively small DNN $T_j(x)$ with a few training epoches.

Specifically, as the support of $\hat{f_{j}}(k)$ is $[\omega_{j}-\frac{\Delta k}{2}, \omega_{j}+\frac{\Delta k}{2}]$, then $\hat{f_{j}}(k-\omega_{j})$\ is supported in
$[-\Delta k/2,\Delta k/2]$, and its inverse Fourier transform $\mathcal{F}^{-1}\left[  \hat{f_{j}}(k-\omega_{j})\right]
,$ denoted as $\ $%
\begin{equation}
f_{j}^{\text{shift}}(x)=\mathcal{F}^{-1}\left[  \hat{f_{j}}(k-\omega
_{j})\right]  (x) \label{rjshift}%
\end{equation}
can be learned quickly by a DNN $T_j(x,\theta)$ by minimizing a loss function
\begin{equation}\label{eq:Lj}
  L_j(\theta) = \int_{-\infty}^{\infty}|f_j^{\text{shift}}(x) - T_j(x,\theta)|^2\d x
\end{equation}
in $n_{0}$-epoches of training.

Moreover, we know that%
\begin{equation}
f_{j}^{\text{shift}}(x_{i})=e^{i\omega_{j}x_{i}}f_{j}(x_{i}),1\leq i\leq N,
\label{dataShift}%
\end{equation}
which provides the training data for $f_{j}^{\text{shift}}(x)$. Equation
(\ref{dataShift}) shows that once $f_{j}^{\text{shift}}(x)$ is learned,
$f_{j}(x)$ is also learned by removing the phase factor.$\mathcal{\ }$%
\begin{equation}
f_{j}(x)\approx e^{-i\omega_{j}x}T_{j}(x,\theta^{(n_{0})}).
\end{equation}
Now with all $f_{j}(x)$ $-M\leq j\leq M$  learned after $n_0$ steps of training each, we have an
approximation to $f(x)$ over all frequency range $[-M\Delta k,M\Delta k]$ as follows%
\begin{equation}
f(x)\approx%
{\displaystyle\sum\limits_{j=-M}^{M}}
e^{-i\omega_{j}x}T_{j}(x,\theta^{(n_{0})}), \label{PhaseDNN}%
\end{equation}
where $\theta^{(n_0)}$ is the value of parameters after $n_0$ steps of training.

\subsection{A parallel phase shift DNN (PhaseDNN) algorithm}

Our goal is to learn a function $f(x)$ using training data%

\begin{equation}
\{x_{i},f_{i}=f(x_{i})\}_{i=1}^{N}. \label{r-trainingdata}%
\end{equation}

In order to apply the decomposition (\ref{x-decomp}) to $f(x)$, when carrying out the sub-training problem \eqref{eq:Lj}, we need to compute the training data for $f_j^{\text{shift}}(x)$ based on original training data \eqref{r-trainingdata}. This  procedure can be done in $x$-space through the
following convolution%

\begin{equation}\label{eq:convol}
\begin{aligned}
f_{j}^{\text{shift}}(x_{i})  &  =e^{i\omega_j x_i}f_{j}(x_{i}) = e^{i\omega_j x_i}\phi_{j}^{\vee}\ast f(x_{i})=\int_{-\infty}^{\infty}\phi
_{j}^{\vee}(x_{i}-s)f(s)ds \\
&  \approx\frac{2\delta}{N_{s}}%
{\displaystyle\sum\limits_{x_{s}\in(x_{i}-\delta,x_{i}+\delta)}}
e^{i\omega_j x_i}\phi_{j}^{\vee}(x_{i}-x_{s})f(x_{s}),
\end{aligned}
\end{equation}
where $\delta$ is chosen such that the kernel function $|\phi^{\vee}(k)|$ is small enough outside $(-\delta, \delta)$.

\section{A coupled PhaseDNN}

\subsection{Approximating functions}

In the previous section, we use the frequency selection kernel $\phi_{j}^{\vee}(x)$ to decompose the training data into different frequency components, each of them after being phase-shifted can be represented by a small DNN. This method can be implemented in parallel. However, it has to use convolution in (\ref{eq:convol})to construct the training data for each small DNN. The convolution is equivalent to a matrix multiplication and the matrix requires a storage of $O(N\times N)$, $N$ is the number of samples. As a result, this convolution process strongly restricts the performance of PhaseDNN for higher dimensions and larger data set.

To avoid this problem, based on the construction of the parallel PhaseDNN (\ref{PhaseDNN}), we would like to consider a coupled weighted phase-shifted
DNNs as an ansatz for a coupled PhaseDNN,
\begin{equation} \label{eq:CPDNNcomplex}%
T(x)=\sum_{m=1}^{M}e^{i\omega_{m}x}T_{m}(x),
\end{equation}
to approximate $f(x),x\in\mathbb{R}^{d}$, where $T_m(x)$ are relatively small complex valued DNNs, i.e., $T_m(x) = T_m^{(real)}(x) + iT_m^{(imag)}(x)$. $T_m^{(real)}(x)$ and $T_m^{(imag)}(x)$ are two independent DNNs. $\{\omega_{m}\}_{m=1}^M$ are frequencies we are particularly interested in from the target function.

We will minimize the following least square loss function
\begin{equation}
\begin{aligned}
L(\theta) &= \int_{-\infty}^{+\infty}|f(x)-T(x)|^2\d x,\label{eq:2}%
\end{aligned}
\end{equation}
or numerically,
%\begin{equation}
%\begin{aligned}
%L_N(\theta) &= \sum_{i=1}^{N}\left\vert
%f(x_{i})-T(x_{i})\right\vert ^{2}\\
%&=\sum_{i=1}^{N}\left\vert f(x_{i})-\sum_{m=1}^{M}e^{i\omega_{m}x_{n}}T_{m}(x_{i})\right\vert ^{2}. \label{eq:Ln2}%
%\end{aligned}
%\end{equation}
\begin{equation}
L_N(\theta) = \sum_{i=1}^{N}\left\vert
f(x_{i})-T(x_{i})\right\vert ^{2}
=\sum_{i=1}^{N}\left\vert f(x_{i})-\sum_{m=1}^{M}e^{i\omega_{m}x_{i}}T_{m}(x_{i})\right\vert ^{2}. \label{eq:Ln2}%
\end{equation}

\begin{Rem}
This method is similar to an expansion with Fourier modes of selected frequency
with variable coefficients defined by DNNs. When $f(x)$ is a real function, it is
equivalent to use real `Fourier' series rather than complex `Fourier' series.
Namely, we will consider the following sine and cosine expansions
\begin{equation}\label{eq:CPDNNreal}%
T(x)=\sum_{m=1}^{M}A_{m}\cos(\omega_{m}x)+B_{m}\sin(\omega_{m}x)
\end{equation}
to approximate $f(x)$, where $A_{m},B_{m}$ are DNNs while $\omega=0$ will
always be included.
\end{Rem}

It can be shown that under the condition that the weights of input layer for each $T_m$ is small, the coupled PhaseDNN is equivalent to the parallel PhaseDNN and a analysis of this fact is given in Appendix. In practical applications, the condition that the weights of input layer for each $T_m$ is small holds at the beginning of training, since we always use small random value to initialize the network. As a matter of fact, to encourage this condition in training process, we can add a weight regularization in the loss function,  namely,
\begin{equation}
L_N^R(\theta) = \sum_{i=1}^{N}|f(x_i) - T(x_i)|^2 + \beta \sum_{m,l} \norm{\vec{W}^{m, l}}_F^2,
\end{equation}
where $x_i$ are training data, $\vec{W}^{m, l}$ is the weight matrix of the $l$-th layer of sub DNN $T_m$, $\beta$ is a regularization parameter. This weight regularization can also restrain some training disasters like gradient blowing up,  etc\cite{deeplearning}.

Comparing with the approach of phase selecting kernel of previous section, the main advantage of the coupled PhaseDNN is that there is
no need for computing convolutions, without the additional quadrature errors, to generate training data for the training of a selected frequency range. This allows us to deal with large data set and higher dimensional problems. However, the coupled PhaseDNN cannot be parallelized and we must choose the  frequencies $\omega_{m}$ before training, then build DNN $T(x)$ using these
$\omega_{m}$ frequencies. If coupled PhaseDNN does not contain enough frequencies, we can modify the coupled phaseDNN with additional frequencies to improve the result.

\subsection{Solving differential equations through least square residual minimization}

The coupled PhaseDNN \eqref{eq:CPDNNcomplex} will be taken as an ansatz for finding the solution of differential equations (DEs) by minimizing the least squares of the DE's residual, similar to the least square finite element (LSFE) method \cite{jiang} \cite{bochev} and the physics-informed neural network (PINN) \cite{raissi2019physics}.

The coupled PhaseDNN will approximate the solution of the following high frequency Helmholtz equation
\begin{equation}%
\mathcal{L}[u]\triangleq u^{\prime\prime}+(\lambda^{2}+c\omega(x))u =f(x),\\
\label{eq:target4}%
\end{equation}
where $\lambda>0$, $c\omega(x)$ can be viewed as a perturbation modeling the inhomogeneity of the otherwise homogeneous media.
% satisfies $\sup_x |c\omega(x)|<\lambda^2$.

The PhaseDNN solution,  in the form of \eqref{eq:CPDNNcomplex} or \eqref{eq:CPDNNreal}, for \eqref{eq:target4} with different boundary conditions can be sought by minimizing the following loss function,
\begin{equation}
L_N(\theta) = L_{ode}(\theta) + \rho L_{bc}(\theta),
\end{equation}
where
\begin{equation}\label{eq:lode}
  L_{ode}(\theta) = \sum_{i=1}^{N}\abs{\mathcal{L}[T](\cdot, \theta)(x_i) - f(x_i)}^2,
\end{equation}
$\{x_i\}_{i=1}^N \in [-1, 1]$ are pre-selected locations to evaluate the residual of the DE by the DNN, and $L_{bc}$ is the boundary condition regularization term, $\rho$ is the regularization parameter.

We consider two typical kinds of boundary value problems. One is Dirichlet boundary condition for an interior Helmholtz problem,
\begin{equation}\label{eq:dirichlet}
  \begin{cases}
    &u^{\prime\prime} + (\lambda^2 + c\omega(x))u = f(x),   \\
    &u(a) = u_1, u(b)=u_2.
  \end{cases}
\end{equation}
For this case, the $L_{bc}$ term is chosen naturally as
\begin{equation}
L_{bc} = (T(a,\theta)-u_1)^2 + (T(b,\theta)-u_2)^2.
\end{equation}

The second type is an outgoing radiation condition for an exterior Helmholtz problem for the wave scattering of
a finite inhomogeneity described by a compact supported function $\omega(x)$,
\begin{equation}\label{eq:scatter}
\begin{cases}
  &u^{\prime\prime} + (\lambda^2 + c\omega(x))u = f(x) \\
  &u^\prime\pm\lambda u\to 0, (x\to \mp\infty).%, \quad u^\prime-\lambda u\to 0, (x\to +\infty).
\end{cases}
\end{equation}

For the exterior problem, we assume both the perturbation $\omega(x)$ and resource function $f(x)$ are compact supported in $[-1,1]$, and we are only interested in the solution in $[-1,1]$. To solve the differential equation on the unbounded domain, we need to truncate the domain to a finite one with an absorbing boundary condition, which in this case is the same as the radiation condition. So, we will consider the following Robin problem of the Helmholtz equation,
\begin{equation}\label{eq:robin}
\begin{cases}
  &u^{\prime\prime} + (\lambda^2 + c\omega(x))u = f(x) \\
  & u^\prime(-a)+\lambda u(-a)=0, \quad u^\prime(a)-\lambda u(a)=0,
\end{cases}
\end{equation}
where a constant $a\geq 2$ is chosen. It can be shown that with $\omega(x)$ and $f(x)$ supported in $[-1,1]$, boundary value problems \eqref{eq:scatter} and \eqref{eq:robin} have the same solution in $[-1, 1]$. The $L_{bc}$ is chosen  as
\[L_{bc} = |T'(-a,\theta)+i\lambda T(-a,\theta)|^2 + |T'(-a,\theta)-i\lambda T(-a,\theta)|^2.\]
Note that the solution is complex valued, $T(x,\theta)$ here should use form \eqref{eq:CPDNNcomplex} and each $T_m$ in \eqref{eq:CPDNNcomplex} should also be complex valued.

\subsection{Solving integral equations for exterior Helmholtz problems}

For exterior scattering problem, a more convenient approach is by converting %
\eqref{eq:scatter} into an integral equation via a Green's function.

When $c=0$, the Green's function of problem \eqref{eq:scatter} is simply
\begin{equation}
G(x,x^{\prime}) = \frac{1}{2i\lambda}e^{i\lambda|x-x^{\prime}|} .
\label{G_ext}
\end{equation}
We can write the solution to \eqref{eq:scatter} with  $c>0$ in terms of $G(x, x^\prime)$ by an
integral equation
\begin{equation}
\begin{aligned} u(x) &= \int_{-\infty}^{\infty} f(x^{\prime})G(x,x^{\prime})
\d x^{\prime} - \int_{-\infty}^{\infty}
c\omega(x^{\prime})u(x^{\prime})G(x,x^{\prime})\d x^{\prime} \\ & =
\int_{-1}^{1} f(x^{\prime})G(x,x^{\prime}) \d x^{\prime} - \int_{-1}^{1}
c\omega(x^{\prime})u(x^{\prime})G(x,x^{\prime})\d x^{\prime} \\ &\triangleq
f_G(x) - \mathcal{K}[u]. \end{aligned}  \label{inteEq}
\end{equation}%
The second equality holds because $f(x)$ and $\omega (x)$ are supported in $%
[-1,1]$. The term $f_{G}(x)$ can be calculated by a Gaussian quadrature before
training.

In order to apply PhaseDNN to approximate the solution of the integral
equation (\ref{inteEq}), we will first discretize the integral operator in a
finite dimensional space by considering a finite element mesh $\{\xi
_{j}\}_{j=1}^{M}$ for the interval $[-1,1]$ and a finite element nodal basis
$\{\phi _{j}(x)\}_{j=1}^{M}$ with the Kronecker property, i,e.,
\begin{equation}
\phi _{j}(\xi _{k})=\delta _{jk}.
\end{equation}

For a function $u(x)$ expressed in term of the basis function $\phi _{j}(x),$
\begin{equation}
u(x)=\sum_{j=1}^{M}u_{j}\phi _{j}(x),\qquad u_{j}=u(\xi _{j}),  \label{eq:expandu}
\end{equation}%
the application of integral operator $\mathcal{K}[u]$ gives

\begin{equation}
\mathcal{K}[u](x)=\sum_{j=1}^{M}u_{j}\int_{-1}^{1}G(x,\xi )\omega (\xi )\phi
_{j}(\xi )d\xi \triangleq \sum_{j=1}^{M}u_{j}\psi _{j}(x),
\end{equation}%
where
\begin{equation}
\psi _{j}(x)=\int_{-1}^{1}G(x,\xi )\omega (\xi )\phi _{j}(\xi )d\xi .
\label{psii(x)}
\end{equation}

Substituting (\ref{eq:expandu}) and (\ref{psii(x)}) into (\ref{inteEq}) , we
have
\begin{equation}
\sum_{j=1}^{M}u_{j}\phi _{j}(x)=f_{G}(x)-c\sum_{j=1}^{M}u_{j}\psi _{j}(x).
\label{eq:expandKu}
\end{equation}%
We will find a DNN $T(x,\theta )$ approximation for solution $u(x)$ by
minimizing the loss function of residual of (\ref{eq:expandKu})\ at $N$%
-locations $\{x_{i}\}_{j=1}^{N}$ with $u_{j}$ replaced by $T(\xi_{j},\theta ),$
\
\begin{equation}
L_{N}(\theta )=\norm{\vec{A}\vec{T}(\theta) + c\vec{B}\vec{T}(\theta) -
\vec{f_G}}^{2},  \label{IntRes}
\end{equation}%
where $\vec{T}(\theta )=[T(\xi _{1},\theta ),T(\xi _{2},\theta ),\dots T(\xi
_{M},\theta )]\in R^{M}$, $\vec{f_{G}}=[(f_{G})(x_{1}),\dots
(f_{G})(x_{N})]\in R^{N}$ and $\vec{A}_{ij}=\phi _{j}(x_{i})$, $\vec{B}%
_{ij}=\psi _{j}(x_{i})$, $1\leq i\leq N,1\leq j\leq M$. The matrix $\vec{B}$
can also be calculated by a Gaussian quadrature before training.

The integral equation method also applies to other type of homogenous
boundary conditions, provided that one can write down the Green's function
for Eqn. \eqref{eq:target4} with the corresponding boundary condition, the
corresponding matrix $\vec{B}$ can be computed by Gaussian quadrature,
similarly.

\begin{Rem}
The integral equation method can also be viewed as a precondition of the least square based method for the differential equation. If we write $\mathcal{L} = \mathcal{L}_1 + c\mathcal{L}_2$, where $\mathcal{L}_1[u] = u'' + \lambda^2u$, $\mathcal{L}_2[u] = \omega(x)u(x)$, the operator $G\ast(\cdot)$  can be regarded as the inverse operator of $\mathcal{L}_1$. Thus the equation \eqref{inteEq} is just $(I + c\mathcal{L}_1^{-1}\mathcal{L}_2)u = \mathcal{L}_1^{-1}f$. When $c$ is small, this precondition is expected to have a better performance than the normal least square based DNN method. This will be confirmed by the numerical results later.
\end{Rem}

\section{Numerical results}

\subsection{Approximation of functions with PhaseDNN}

\subsubsection{Parallel PhaseDNN}

In this section, we will present numerical results to demonstrate the
capability of PhaseDNN to learn high frequency content of target functions. In
practice, we could sweep over all frequency ranges with a prescribed frequent increment $\Delta
k=5$. For the test function for which we
have some rough idea about the the range of frequencies in the data, only a few
frequency intervals are selected for the phase shift.

We choose a target function $f(x)$ in $[-\pi, \pi]$
\begin{equation}
\label{eq:target1}f(x) =
\begin{cases}
10(\sin x + \sin3x), & \mbox{if } x\in[-\pi,0],\\
10(\sin23x + \sin137x + \sin203x), & \mbox{if }x\in[0,\pi].
\end{cases}
\end{equation}

\begin{figure}[ptbh]
\centering  \includegraphics[scale = 0.5, angle=90]{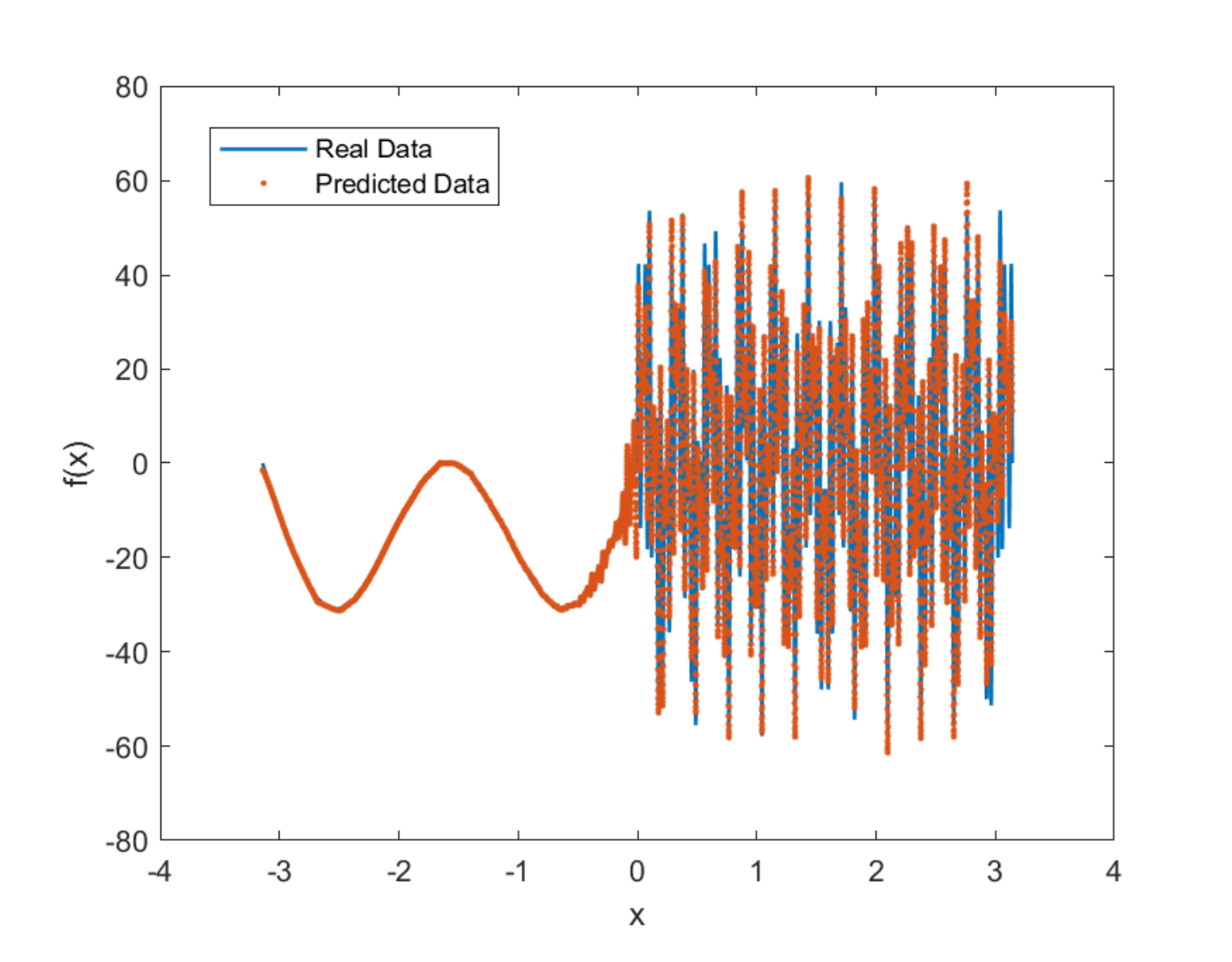} \caption{The
testing result of $f_{DNN}(x)$ trained by the parallel PhaseDNN. The blue solid line is
$f(x)$ and the data marked by red dots are the value of $f_{DNN}(x)$ at
testing data set.}%
\label{fig:fivefreq}%
\end{figure}

\begin{figure}[ptb]
\centering \scalebox{0.4}{\includegraphics[angle=90]{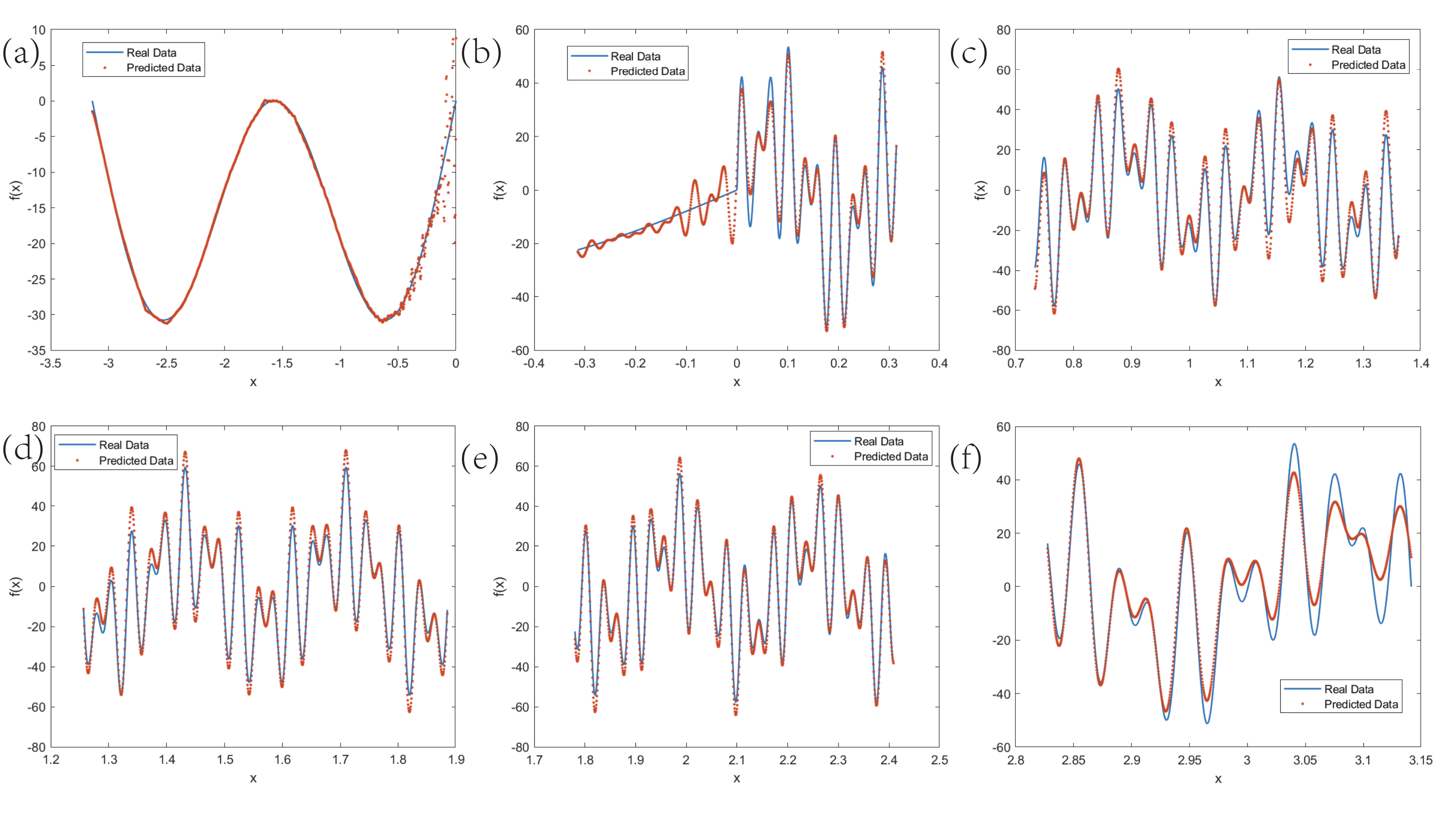}}
\caption{The detail results of training in different intervals. The subfigures
(a)-(f) shows the results in interval $[-\pi,0]$, $[-\pi/10,\pi/10]$,
$[\pi/3-\pi/10,\pi/3+\pi/10]$, $[\pi/2-\pi/10,\pi/2+\pi/10]$, $[2\pi
/3-\pi/10,2\pi/3+\pi/10]$ and $[\pi-\pi/10,\pi]$ correspondingly. The blue
solid line is $f(x)$ and the data marked by red dots are the value of
$f_{DNN}(x)$ at testing data set.}%
\label{fig:fivefreqdetail}%
\end{figure}

Because the frequencies of this function is well separated, we need not to
sweep all the frequencies in $[-\infty, +\infty]$. Instead, we choose $\Delta
k=5$, and use the following functions
\[
\begin{aligned}
&\phi_1(k) = \chi_{[-205,-200]}(k)  & \phi_2(k) = \chi_{[-140,-135]}(k)\\
& \phi_3(k) = \chi_{[-25,-20]}(k)  & \phi_4(k) = \chi_{[-5,0]}(k)\\
&\phi_5(k) = \chi_{[0,5]}(k)   & \phi_6(k) = \chi_{[20,25]}(k)\\
& \phi_7(k) = \chi_{[135,140]}(k)  & \phi_8(k) = \chi_{[200,205]}(k)
\end{aligned}
\]
to collect the frequency information in the corresponding frequency intervals
and shift the center of the interval to the origin by a phase factor. For each
$f_{j}(x) = \mathcal{F}^{-1}[\hat{f}\phi_{j}](x)$, we construct two DNNs
to learn its real part and imaginary part, separately. Every DNN have 4 hidden
layers and each layer has 40 neurons. Namely, the DNN has a structure
1-40-40-40-40-1. The training data is obtained by 10,000 samples from the
uniform distribution on $[-\pi,\pi]$ and the testing data is 10000 {\color{black} evenly spaced points in $[-\pi,\pi]$}. We train these DNNs with 1000 epochs by Adam optimizer with training rate
0.002 and batchsize 2000 for each DNN. The result is shown in Fig.
\ref{fig:fivefreq} while the detail of the training result is shown in Fig. \ref{fig:fivefreqdetail}.
These figures clearly shows that phase DNN can capture the various high
frequencies, from low frequency $\pm1$, $\pm3$ to high frequency $\pm203$
quite well. The training time of PhaseDNN are collected in Table
\ref{tab:phaseDNN}

\begin{table}[ptbh]
\centering
\begin{tabular}
[c]{|c|c|c|c|c|}\hline
Frequency  & Convolution & Training & Total Time\\
Interval & time(s) & Time(s) &\, (s) \\
\hline
$[-205,-200]$ & 11.32 & 15.05 & 26.38\\
$[-140,-135]$ & 11.32 & 15.18 & 26.51\\
$[-25,-20]$ & 11.46 & 14.99 & 26.45\\
$[-5,0]$ & 10.70 & 14.97 & 25.67\\
$[0,5]$ & 11.22 & 14.98 & 26.21\\
$[20,25]$ & 11.26 & 15.00 & 26.27\\
$[135,140]$ & 11.32 & 15.13 & 26.45\\
$[200,205]$ & 11.32 & 15.03 & 26.36\\
\hline
Total & 89.94 & 120.37 & 210.32\\
\hline
\end{tabular}
\caption{The training time statistics. For each $j$, the training
time is the sum of training time of real and imaginary part. Each DNN is trained by 1000 epochs with batchsize 2000.}%
\label{tab:phaseDNN}%
\end{table}

It is shown that the convolution calculus for preparing data for $f^{\rm shift}_j(x)$ costs about 40\% of the total computing time. It is quite a large portion and inefficient. Because in different interval, $f_{j}(x)$ can be trained in parallel, PhaseDNN is ideal to take advantage of parallel computing architectures. Although the total computing time is 210 seconds, in practice, the computation can be done in 27 seconds with parallelization.
%In comparison with one single normal DNN, we have used a 24 hidden layers with 640
%neurons per hidden layer with the same amount of training data, the loss is still around 100 after more than 50000 epoch of training taking over 5 hours nonstop running on the same workstation, see the following Fig.\ref{Fig:1}(c).
In comparison, a normal single fully connected 24 layer DNN with 640 neurons per hidden layer shows non-convergence in Fig.\ref{Fig:1}(c) and (d) after over 5 hours of training.

\subsubsection{Coupled PhaseDNN}

\textbf{1-D\ Problem:} We will apply the coupled $\quad$ PhaseDNN method to the same test problem \eqref{eq:target1}.
The frequencies $\{\omega_{m}\}$ are selected to be $0,5,25,135,200$. For each $A_{m}$ and $B_{m}$, we also set it as a
1-40-40-40-40-1 DNN.

\begin{figure}[ptbh]
\centering  \includegraphics[angle=90, width=12cm]{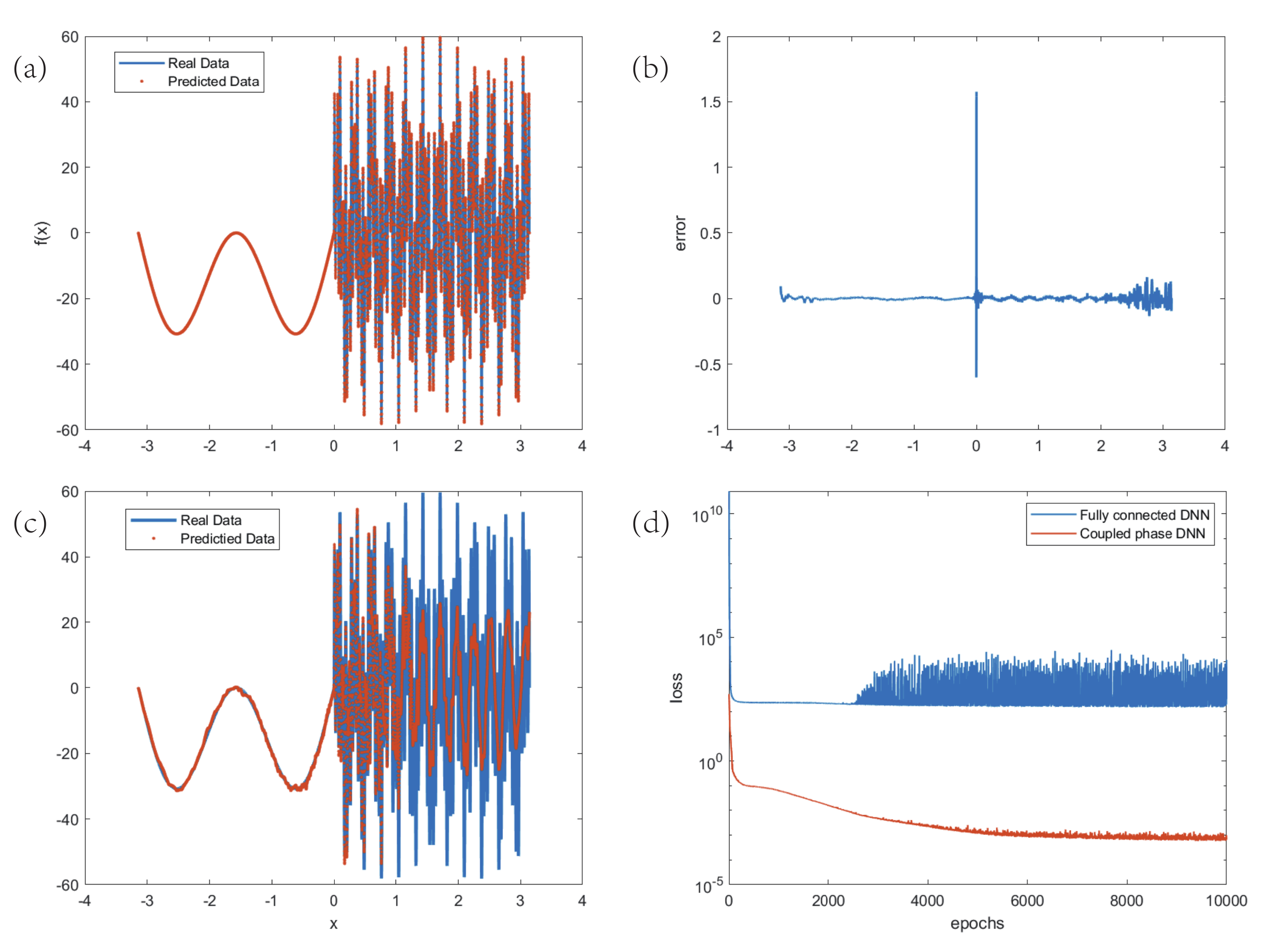}
\caption{Fitting result for $f(x)$ using coupled PhaseDNN can a single fully connected DNN. (a) Fitting result of couple phase DNN after 10000 epochs of training. We select frequency $\{\omega_{m}\} = \{0,5,25,135,200\}$. The blue solid line is real data while the red dots are predicted value of couple PhaseDNN. (b) The pointwise error of coupled PhaseDNN. (c) Fitting result of a fully connected DNN after 50000 epochs of training. The DNN has 24 layers and 640 neurons in each layer. (d) The convergence properties of coupled phase DNN and single fully connected DNN in log scale. The blue line is training error of fully connected DNN. The red line is training error of coupled phase DNN.}%
\label{Fig:1}%
\end{figure}

The training parameters are set as the same as before. Testing data is 10000 {\color{black} evenly spaced points in $[-\pi,\pi]$}. The testing result is shown in Fig.\ref{Fig:1}(a). The average $L^{2}$ relative training error and testing error are both $1.4\times10^{-3}$.  The pointwise testing error is shown in Fig\ref{Fig:1}(b). It is clear that the error is concentrated in the neighborhood of 0, where the derivative
of $f(x)$ is discontinuous. Out of this neighborhood, the relative maximum error is $8\times 10^{-3}$.

To show the accuracy and efficiency of the coupled PhaseDNN, we try to learn $f(x)$ by a single fully connected DNN. The DNN is set to have 24 hidden layers and 640 neurons in each layer. The training is also carried out with 10000 random training samples, 2000 batchsize and learning rate $0.001$. After 50000 epochs during 5 hours of training, the result with a total loss of 100 is shown in Fig.\ref{Fig:1}(c) and (d) ({\color{black} blue} line).

One can see that a single fully connected DNN cannot learn this highly oscillated function even with such a large network and a very long training time. The convergence behavior of a single DNN and coupled PhaseDNN are shown in Fig.\ref{Fig:1}(d). It is shown that the training loss of coupled phaseDNN reduces quickly to $O(10^{-1})$ after 1000 epochs while the loss of a single DNN stays $O(10^2)$ even after 10000 epochs. Coupled PhaseDNN is proved efficient in learning high frequency functions.

%We can also sweep the frequency space by choosing the frequencies $w_{m}$ to
%be -250:10:250, training parameters are the same as before. The fitting result
%and error is shown in Fig.\ref{Fig:2}. The average $L^{2}$ training and
%testing error are reduced to $4.2\times10^{-4}$.
%
%\begin{figure}[ptbh]
%\centering  \includegraphics[angle=90, width=12cm]{1D-sweep.pdf}
%\caption{Fitting result for $f(x)$ using sweep method. Frequency domain is
%[-250,250]. Left panel: fitting. Right panel: error.Fitting result for $f(x)$
%using sweep method. Frequency domain is [-250,250]. Left panel: fitting. Right
%panel: error.}%
%\label{Fig:2}%
%\end{figure}
%
%For comparison, we can also use a fully connected DNN with similar scale as
%coupled phaseDNN to learn $f(x)$. This DNN has 4 hidden layers with 360 neuron
%in each layer. This DNN was trained with the same parameter as before. The
%result is shown in the right panel of Fig.\ref{Fig:1}. It is clear that almost
%nothing about the high frequency information was learned by this fully connected DNN..

\begin{figure}[ptbh]
\centering  \includegraphics[angle=90, width=12cm]{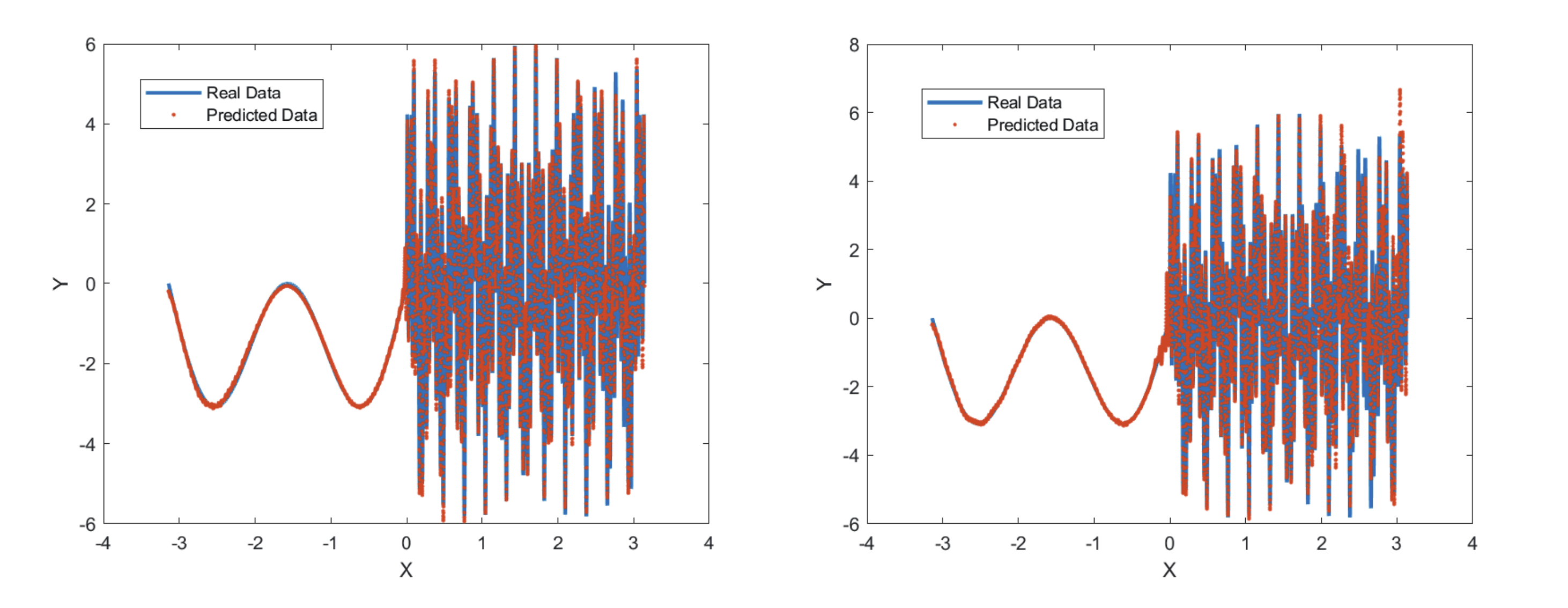}
\caption{Fitting result for $f(x)$ using less data. We select frequency
$w_{n}\in\{0,5,25,135,200\}$. Left panel: training with 1000 random samples.
Right panel: training with 500 random samples.}%
\label{Fig:3}%
\end{figure}

In fact, 10000 samples are too much for this example. It turns out that 1000
samples can lead to a good approximation with Eqn. \eqref{eq:target1}. Even with 500
samples, which is a much too small data set for the frequency 203, We can
still get a `reasonable' result. The testing results with 10000 evenly spaced points in $[-\pi,\pi]$ is
shown in Fig.\ref{Fig:3}.

\medskip

\begin{itemize}
\item \noindent\textbf{Discontinuous functions and frequency sweep}
\end{itemize}

\begin{figure}[ptbh]
\centering  \includegraphics[angle=90, width=12cm]{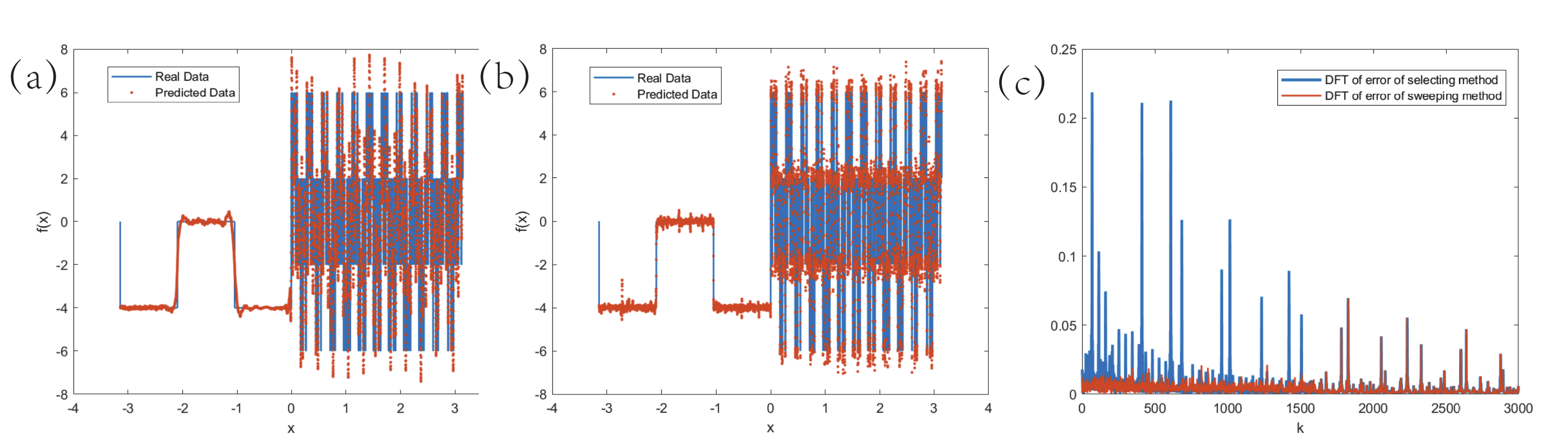}
\caption{Fitting result for square wave function using selecting and sweeping
methods. (a) Fitting result using selecting method with $\omega_{m}\in\{0, 5, 25, 135, 200\}$. (b) Fitting result using sweeping method. Frequency domain is [-1600,1600]. (c) DFT of error of selecting and sweeping methods. The blue line is the DFT of error of selecting method. The red line is DFT of error of sweeping method.}%
\label{Fig:4}%
\end{figure}

Next we consider discontinuous functions and we replace the $\sin$ in
Eqn.\eqref{eq:target1} by square wave function with same frequency and learn
it by \eqref{eq:CPDNNreal} with $\omega_{m}\in\{0, 5, 25, 135, 200\}$ and $\omega_{m}=-1600:10:1600$.
These two DNNs are trained with 10000 samples and 1000 epochs. The results are
shown in Fig.\ref{Fig:4}(a), (b). It can be seen that the coupled PhaseDNN has a larger error for discontinuous function, compared with the case for the smooth $\sin$ case. The sweeping strategy is preferred for discontinuous functions. From the plot of error's DFT in Fig.\ref{Fig:4}(c), one may find that the sweeping strategy does learns the information in frequency domain $[-1600, 1600]$. For the frequency larger than 1600, neither strategy can learn it.

\begin{figure}[ptbh]
\begin{center}
\includegraphics[width=4.5in]{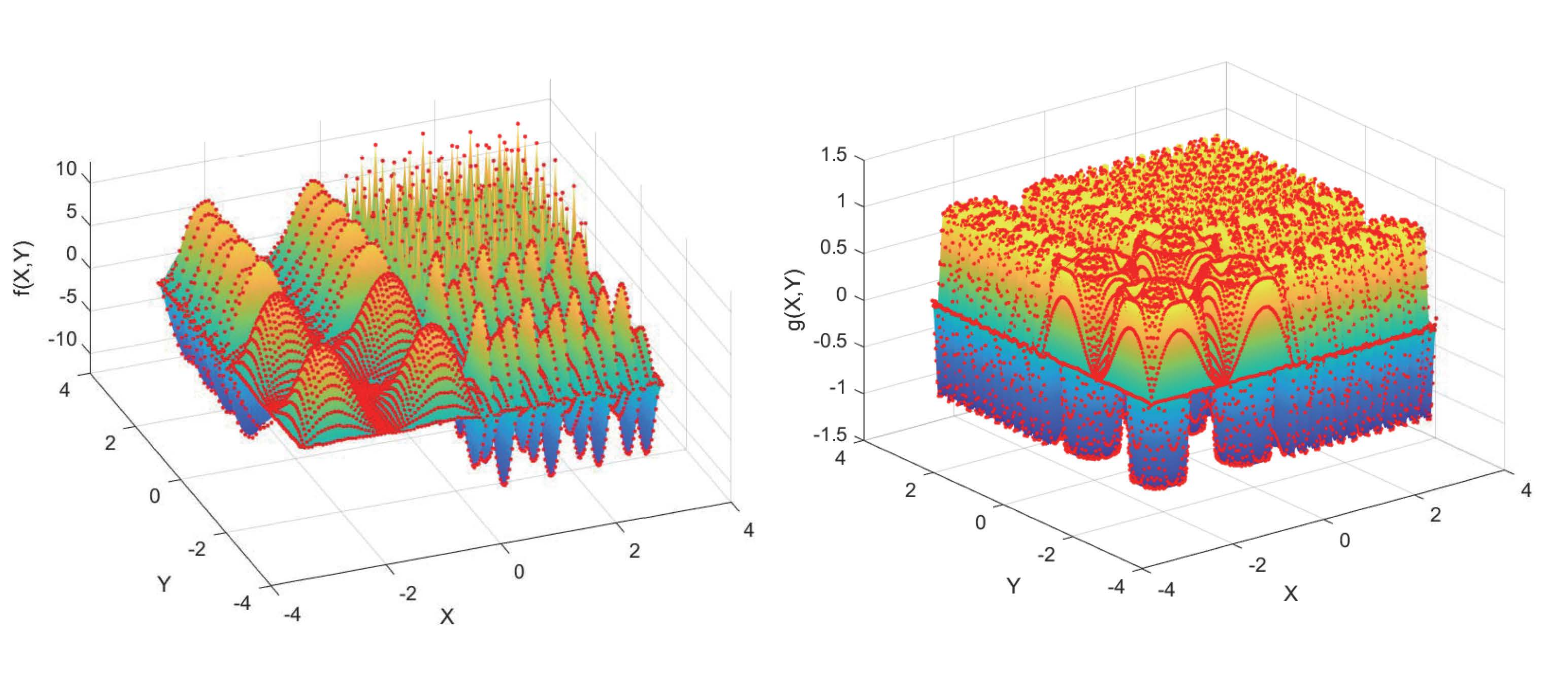} 
\end{center}
\caption{Fitting results for 2D problems using coupled PhaseDNN. Left panel: fitting result for
$G(x,y)$. Right panel: fitting result for $\tilde{G}(x,y)$.}%
\label{Fig:5}%
\end{figure}

\textbf{2-D\ problem:} We use Eqn. \eqref{eq:CPDNNreal} to learn 2D and 3D problems.
For 2D test, the function $G(x,y)=g(x)g(y)$ is used, where $g(x)$ is
defined by%
\begin{equation}
\label{eq:target2}g(x) =
\begin{cases}
\sin x + \sin3x, & \mbox{if } x\in[-\pi,0],\\
\sin23x + \sin137x , & \mbox{if }x\in[0,\pi].
\end{cases}
\end{equation}
The function $g(x)$ is the $f(x)$ in \eqref{eq:target1} without the $\sin203x$
component.
%In our precious 2D test, We can only deal with about 20000 samples
%due to the storage constrain of convolution. Now, because there is no
%convolution any more, we can deal more data.
In this test, we choose $\{w_{m}\}\in\{0,5,25,135\}\times\{0,5,25,135\}$. Training
setting are $640\times640=409600$ samples and 80 epochs with batchsize 100.
Testing data is $100\times100$. The result is shown in Fig.\ref{Fig:5} (left).
%
%\begin{figure}[ptbh]
%\centering  \includegraphics[width=12cm]{2D.pdf} \caption{Fitting result for
%$G(x,y)$.}%
%\label{Fig:5}%
%\end{figure}

The result is good even for the highest frequency region.
%In our
%previous 2D test, we can only deal with a frequency about 30.
In this example, the highest frequency is 137. With more data, we can learn a
function of even higher frequency.

Furthermore, we test another problem with $\tilde{G}(x,y)=\sin(\tilde
{g}(x)\tilde{g}(y))$ with%

\begin{equation}
\label{eq:target22}\tilde{g}(x) =
\begin{cases}
\sin x + \sin3x, & \mbox{if } x\in[-\pi,0],\\
\sin23x , & \mbox{if }x\in[0,\pi].
\end{cases}
\end{equation}

The highest frequency of $\tilde{G}(x,y)$ is about 200. We use the similar
training setups as the previous test and choose $\omega_{m}\in\{10:10:210\}\times
\{10:10:210\}$. The fitting result is shown in  Fig.\ref{Fig:5} (right). The $L^{2}$
fitting error is $5.2\times10^{-3}$.

%\begin{figure}[ptbh]
%\centering  \includegraphics[width=12cm]{fitting_sin.pdf} \caption{Fitting
%result for $\tilde{G}(x,y)$.}%
%\label{Fig:11}%
%\end{figure}

%\textbf{3-D Problem:} The test problem for 3D is $H(x,y,z)=h(x)h(y)h(z)$,
%where
%\begin{equation}
%h(x)=%
%\begin{cases}
%\sin x+\sin3x, & \mbox{if }x\in\lbrack-\pi,0]\\
%\sin23x+\sin32x, & \mbox{if }x\in\lbrack0,\pi].
%\end{cases}
%\label{eq:target3}%
%\end{equation}
%The selected frequency is $\omega_{m}\in\{0,25,30\}\times\{0,25,30\}\times
%\{0,25,30\}$. Training uses $150\times150\times150=3375000$ random samples and
%50 epoches with batchsize 1000. For plotting, we
%choose hypersurface $z=1$ and $x+y+z=1$ as test data. Each $A_{m},B_{m}$ is
%chosen to be 1-20-20-20-20-1 DNN. The whole network is trained 40 epochs with
%batchsize 1000.  {\color{black} The relative $L^2$ error is 0.4.} Results are shown in Fig.\ref{Fig:6}.
%
%\begin{figure}[ptbh]
%\centering  \includegraphics[angle=90, width=12cm]{3D.pdf} \caption{Fitting
%result for $H(x,y,z)$. Left panel: fitting result on hypersurface $z=1$. Right
%panel: fitting result on hypersurface $x+y+z=1$.}%
%\label{Fig:6}%
%\end{figure}

\begin{Rem}
\textbf{The number of data}. Basically,the data set must be big enough so that
it can reveal all the frequencies. That means we still need $O(N^{d})$ data.
For each direction, $N$ samples must reveal the highest frequency of this
direction. This means, even though DNN has the advantage that the number of unknowns $p$
increases linearly w.r.t the number of dimension, we still need an exponentially large data set. In our 3D
test, 150 random samples for each direction is merely enough for frequency 32,
the total number $3.4\times10^{6}$ is already a very big data set.
\end{Rem}

It is clear that the PhaseDNN approach cannot overcome curse of dimensionality. If we have no prior knowledge on the frequency distribution, coupled phaseDNN can be considered as building a mesh in Fourier space. Thus, in general, the number of $\omega_{m}$ will increase exponentially. In our 3D
test problem, there are 108 different $\omega_{m}$, which corresponds to 216 sub
DNNs. The whole coupled phaseDNN $T(x)$ has a width over 4000. It is a shallow but very wide DNN. The number of parameters is large.  With a large number of data, the whole training takes 5 hours.

\subsection{Coupled PhaseDNN for solving high frequency wave problems}

\subsubsection{Helmholtz equation with constant wave numbers}

\begin{figure}[ptbh]
\centering  \includegraphics[angle=90, width=12cm]{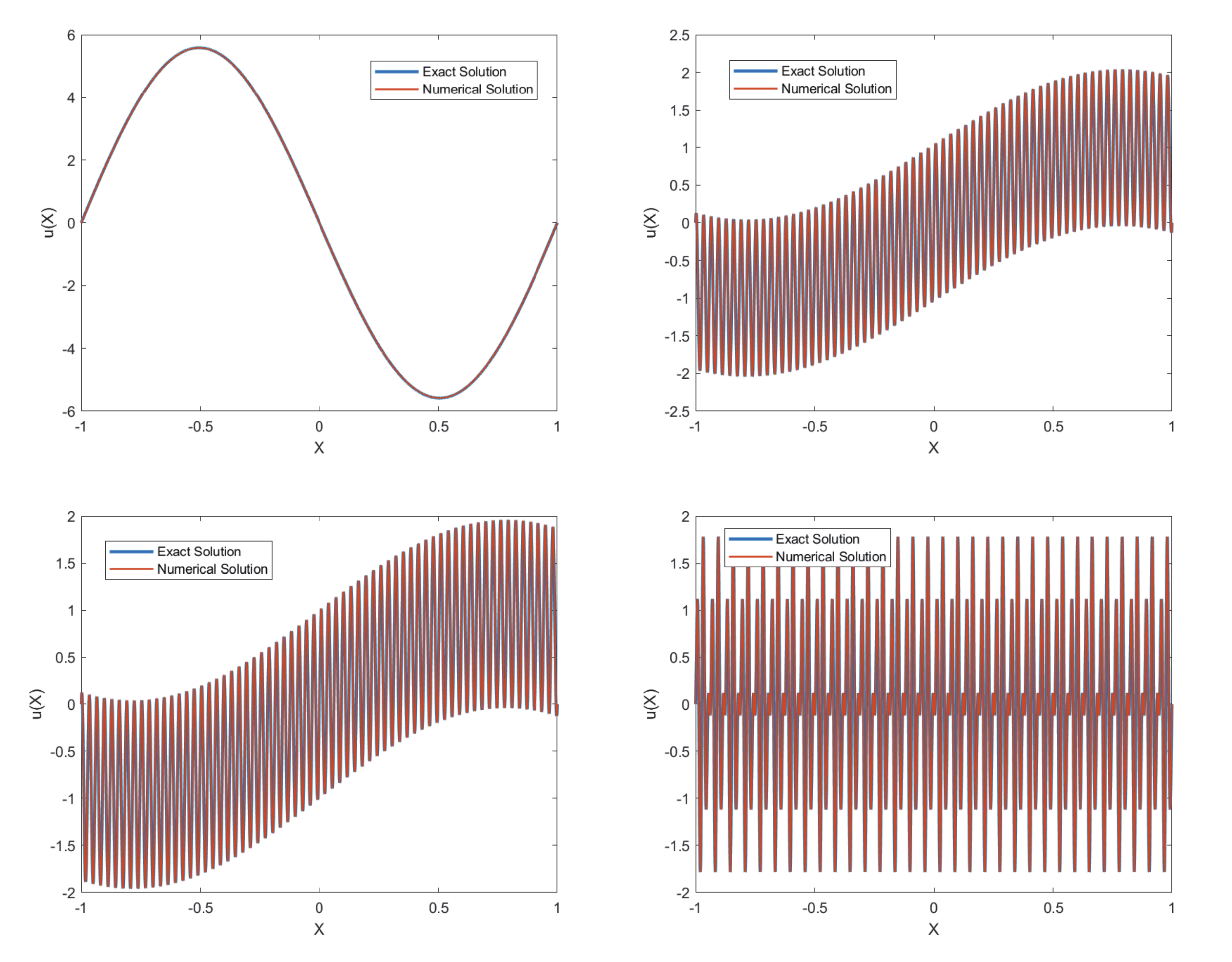} \caption{Numerical
and exact solution of problem \eqref{eq:dirichlet} with $c=0$ and different $\lambda$ and
$\mu$. Upper left panel: $\lambda=3, \mu=2$. Upper right panel: $\lambda=200,
\mu=2$. Lower left panel: $\lambda=2, \mu=200$. Lower right panel:
$\lambda=300, \mu=200$.}%
\label{Fig:7}%
\end{figure}

We solve problem \eqref{eq:dirichlet} with $c=0$ and boundary condition $u(-1)=u(1)=0$. We set $\omega_{m}\in\{0, \lambda, \mu\}$, each $A_{m}, B_{m}$ to be
1-40-40-40-40-1 DNN. The whole $T(x, \theta)$ is trained with 10000 evenly spaced samples, 100 epochs and batchsize 100. We choose four special cases:
$\lambda=3, \mu=2$; $\lambda=200, \mu=2$; $\lambda=2, \mu= 200$; and $\lambda=300, \mu=200$. The result is shown in Fig.\ref{Fig:7}.

\begin{figure}[ptbh]
\centering  \includegraphics[angle=90, width=12cm]{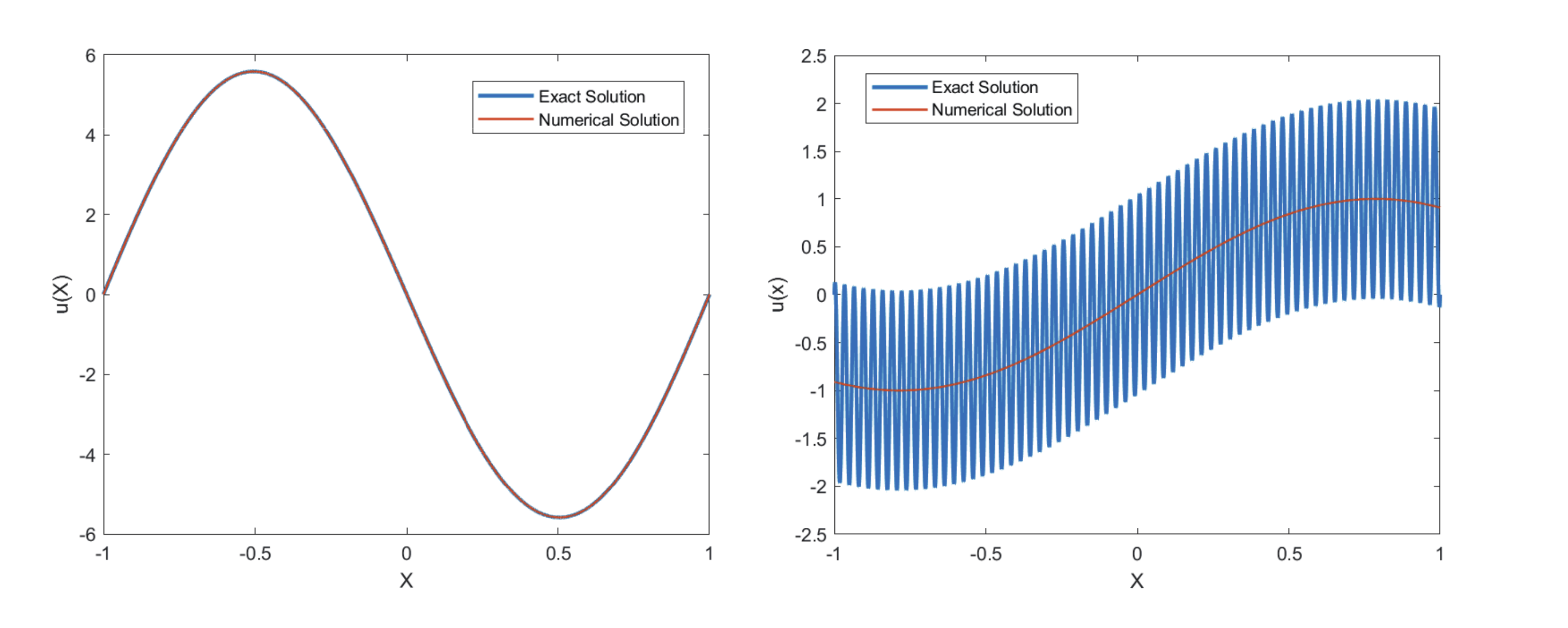}
\caption{Non-convergence of usual fully connected DNN for high frequency case:
DNN and exact solution of problem \eqref{eq:dirichlet} with different $\lambda$ and
$\mu$. Left panel: $\lambda=3,\mu=2$. Right panel: $\lambda=200,\mu=2$. }%
\label{Fig:8}%
\end{figure}

The training takes about 5 minute with a maximum error is $O(10^{-4})$.
 For comparison, a single fully connected DNN with
similar scale as $T(x)$ cannot solve the equation at all when the frequency is
high. The training result after 1500 epochs for $\lambda=3, \mu=2$ and
$\lambda=200, \mu=2$ are shown in Fig.\ref{Fig:8}, showing the non-convergence for high frequency solution using a common fully connected DNN (Fig.\ref{Fig:8} (right)).

\subsubsection{Helmholtz equation with variable wave numbers}

Next we solve problem \eqref{eq:dirichlet} with $u(-1)=u(1)=0$ and $c>0$, and a variable wave number $\omega(x) = \sin(mx^2)$, $m>0$ is a constant. As there is no explicit exact solution to this equation, the numerical
solution obtained by a finite difference method with a fine mesh will be used as the reference solution.
%
%We first choose $\lambda=100$, $\mu=10$, $c=10$ and $m=1$ in
%Eqn.\eqref{eq:ode_vc}, which corresponds to a low frequency external wave
%source and a low wave number with small background media inhomogeneity. In the
%coupled phaseDNN, we choose $\omega_{m}\in\{0, 10:10:150\}$. Other training
%parameters are set to be similar as in the constant coefficient case. The
%numerical result of coupled phaseDNN and reference solution is shown in
%Fig\ref{Fig:9a} and the absolute error is in the order of $O(10^{-3})$.
%
%\begin{figure}[ptbh]
%\centering  \includegraphics[angle=90, width=12cm]{VC_high_lambda_low_mu_sinx.pdf}
%\caption{Numerical and exact solution of Eqn.\eqref{eq:ode_vc} using coupled
%phaseDNN $\lambda=2$, $\mu=10$, $c=3.6$ and $m=1$. Left panel: The numerical
%solution and reference solution obtained by finite difference method. Right
%panel: The absolute value of the difference between numerical solution and
%reference solution. }%
%\label{Fig:9a}%
%\end{figure}

\begin{figure}[ptbh]
\centering  \includegraphics[angle=90, width=12cm]{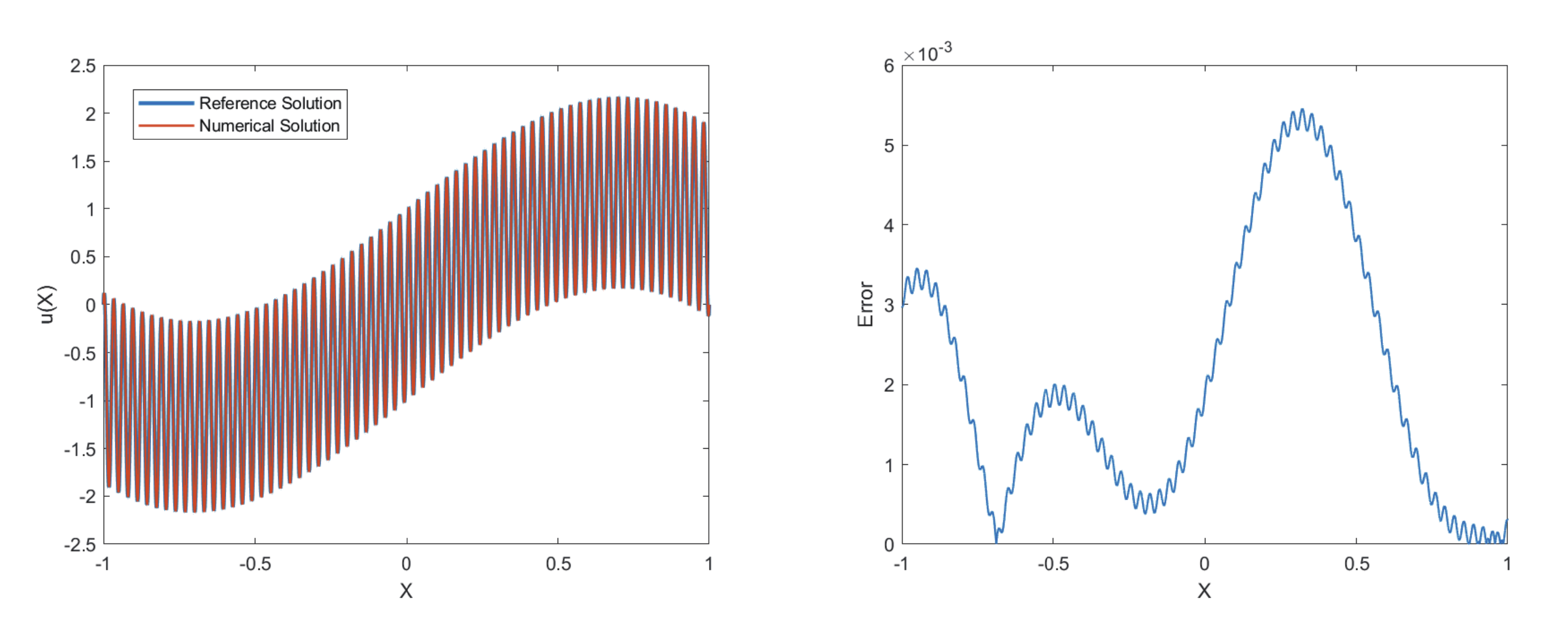}
\caption{Numerical and exact solution of problem \eqref{eq:dirichlet} using coupled
phaseDNN. $\lambda=2$, $\mu=200$, $c=3.6$ and $m=1$. Left panel: The numerical
solution and reference solution obtained by finite difference method. Right
panel: The absolute value of the difference between numerical solution and
reference solution. }%
\label{Fig:9}%
\end{figure}

We first choose $\lambda=2$, $\mu=200$, $c=0.9\lambda^{2}=3.6$ and $m=1$ in Eqn.\eqref{eq:dirichlet}, which corresponds to a high frequency external wave source and a low wave number with small background media inhomogeneity. In the coupled PhaseDNN, we choose $w_{m}\in\{1,2,3,4,200\}$. Other training
parameters are set to be similar as in the constant coefficient case. The numerical result of coupled phaseDNN and reference solution is shown in Fig. \ref{Fig:9} and the absolute error is in the order of $O(10^{-3})$.

\begin{figure}[ptbh]
\centering  \includegraphics[angle=90, width=12cm]{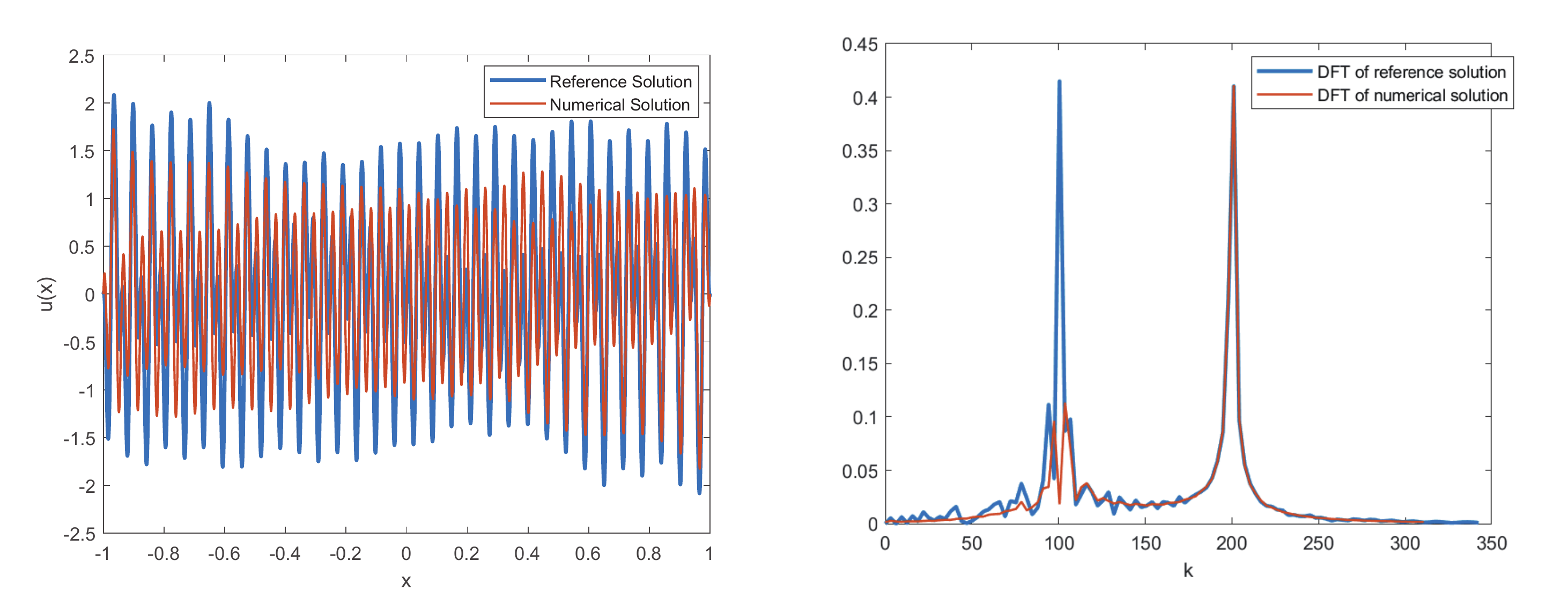}
\caption{Numerical and reference solutions to problem \eqref{eq:dirichlet} using coupled
PhaseDNN. $\lambda=100$, $\mu=200$, $c=0.1\lambda^2$ and $m=100$. Left panel: The numerical
solution obtained by least square based coupled PhaseDNN method and the reference solution. Right panel: the discrete fourier transform of reference solution and numerical solution. Blue line: DFT of reference solution. Red line: DFT of PhaseDNN numerical solution.}%
\label{Fig:10}%
\end{figure}

Next, we choose $\lambda=100$, $\mu=200$, $c=0.1\lambda^2=1000$ and $m=100$, which
corresponds to a high frequency external wave source and a high wave number
with larger background media inhomogeneity. $\omega_{m}$ is chosen to be
$\{0,90,100,110,190,200,210\}$. The learning result is shown in Fig. \ref{Fig:10}.

One can see that {\color{black} coupled PhaseDNN} with least square residual minimization cannot solve the problem well.
The right panel of Fig.\ref{Fig:10} shows the error concentrates on the $\pm\lambda$ component while the $\mu$ component converges quite good. The error of this method is mainly due to high wave number. This is an intrinsic difficulty of {\color{black} least square minimization based DNN method}. The theoretical analysis of this phenomena will be done in a coming paper.

Next, we will apply the integral equation approach (\ref{inteEq}) (\ref{IntRes}) to this problem. The Green's function to $u''+\lambda^2u=\delta(x-x')$ with Dirichlet boundary condition $u(-1)=u(1)=0$ is given by
\begin{equation}
G(x{^{\prime}},x)=\left\{
\begin{aligned} &\frac{ (-\tan{\lambda} \cos{\lambda x^{'}}+\sin{\lambda x^{'}})(\tan{\lambda}\cos{\lambda x}+\sin{\lambda x})}{2\lambda \tan{\lambda}}, s \leq x,\\ &\frac{ (\tan{\lambda} \cos{\lambda x^{'}}+\sin{\lambda x^{'}})(-\tan{\lambda}\cos{\lambda x}+\sin{\lambda x})}{2\lambda \tan{\lambda}}, s > x \end{aligned}\right.
\label{G_int}%
\end{equation}

\begin{figure}[ht]
    \includegraphics[width=0.48\linewidth]{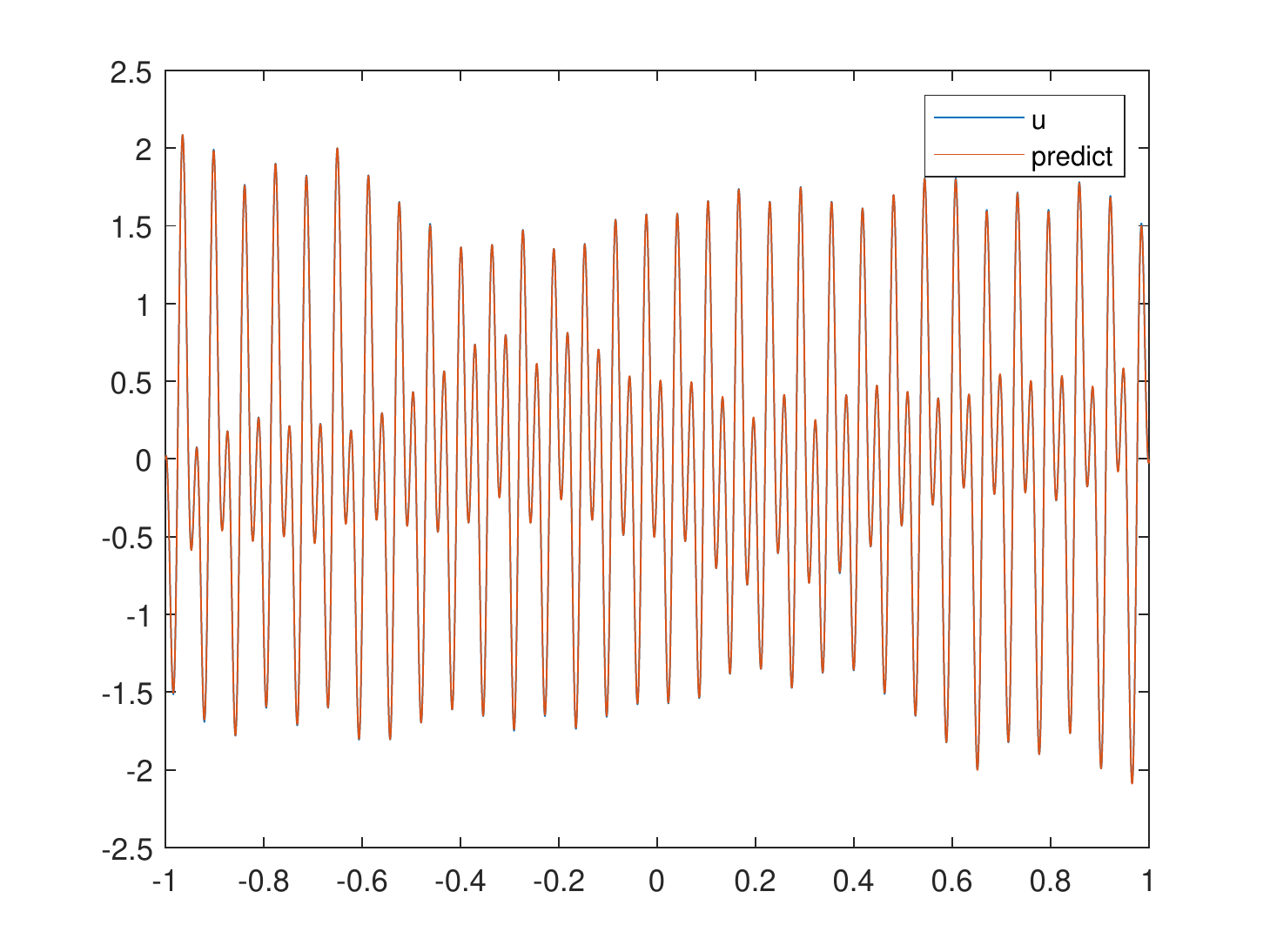}
    \includegraphics[width=0.48\linewidth]{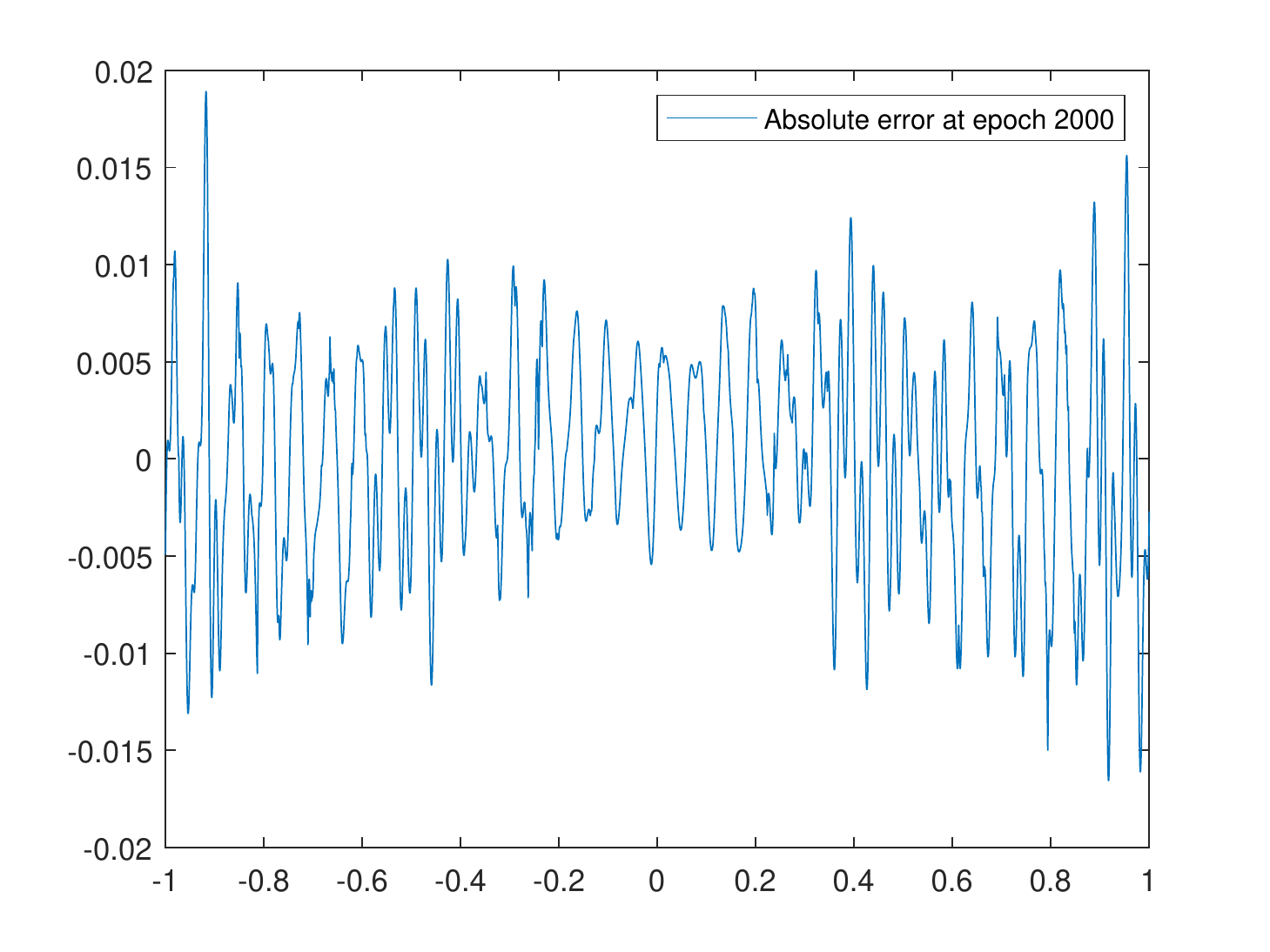}
    \caption{Numerical and reference solution of problem \eqref{eq:dirichlet} using coupled
    phaseDNN {\color{black} and integral equation method}. $\lambda=100$, $\mu=200$, $c=1000$ and $m=100$. Left panel: The numerical
    solution with reference solution. Right panel: The difference between numerical solution and reference solution. }
    \label{fig:6}
\end{figure}

With the same parameter setting, the numerical solution obtained by integral equation method is shown in Fig.\ref{fig:6}. The absolute error is in the order of $O(10^{-3})$.

It can be seen that coupled PhaseDNN with the integral equation approach (\ref{inteEq}) (\ref{IntRes}) gives more accurate solution than that for the differential equation method and does not suffer from high wave number errors as in Fig.\ref{fig:6}.

\subsubsection{Solving elliptic equation}

We can also solve elliptic differential equation with high frequency external sources using the coupled PhaseDNN. We consider a test problem
\begin{equation}
\label{eq:elliptic}%
\begin{cases}
u^{\prime\prime}- \lambda^{2}u = -(\lambda^{2} + \mu^{2})\sin(\mu x),\\
u(-1) = u(1) = 0,
\end{cases}
\end{equation}
which has an exact solution as
\begin{equation}
\label{eq:exactellip}u(x) = -\frac{\sin\mu}{\sinh\lambda}\sinh(\lambda x) +
\sin(\mu x).
\end{equation}

\begin{figure}[ptbh]
\centering  \includegraphics[angle=90, width=12cm]{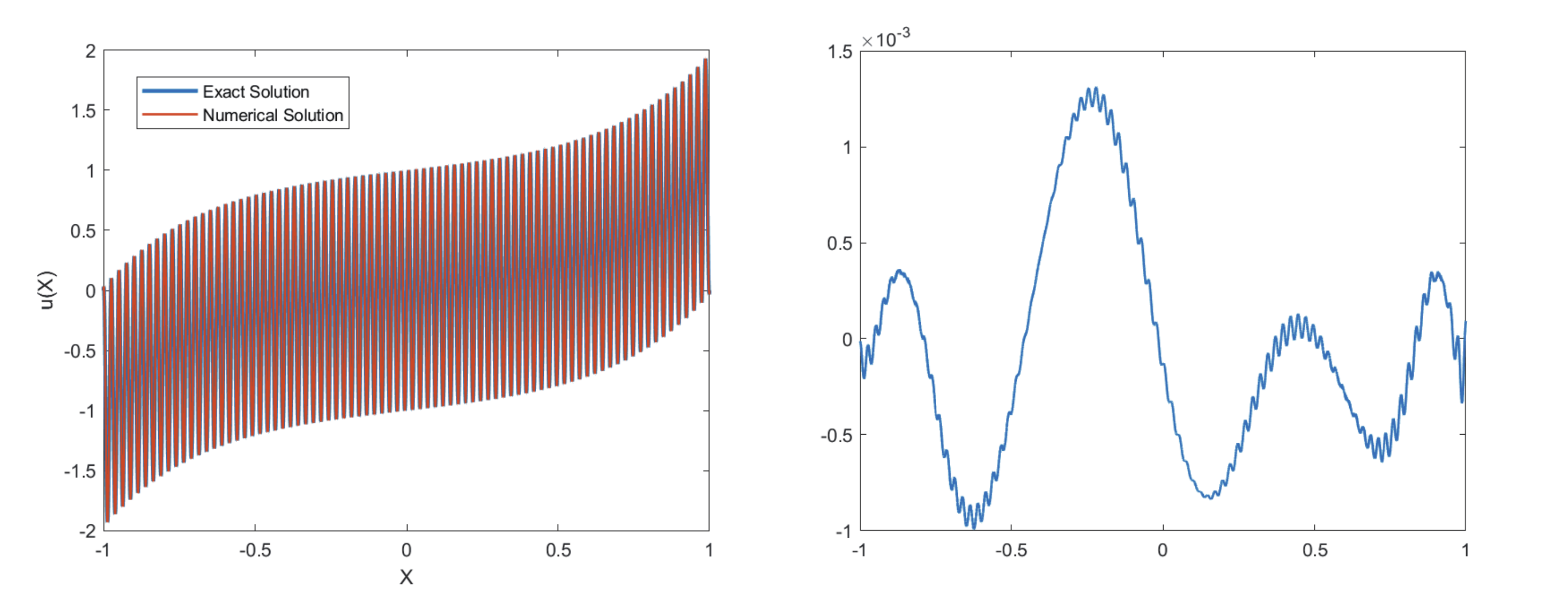} \caption{The
numerical solution of equation \eqref{eq:elliptic} using coupled PhaseDNN.
$\lambda=3$, $\mu=250$. Left panel: Numerical solution(in red) and exact
solution(in blue) of equation \eqref{eq:elliptic}. Right panel: the error of
numerical solution. }%
\label{Fig:12}%
\end{figure}

We choose $\lambda=3$, $\mu=250$ in Eqn. \eqref{eq:elliptic}. To solve this
equation, we set a coupled PhaseDNN with $\omega_{m}\in\{0, \mu\}$. Each
subnetwork is a fully connected DNN with 4 layers and 20 neurons in each
layer. Accurate training result is shown in Fig. \ref{Fig:12}.

\subsubsection{Coupled PhaseDNN for solving exterior wave scattering problem}

We consider problem \eqref{eq:scatter} with $\lambda = 100$, $\mu=200$, $c=0.1\lambda^2$. The variable coefficient $\omega(x) = \chi_{[-1,1]}(x)\sin(1-x^2)$, and  $f(x) = \chi_{[-1,1]}(x)(\lambda^2-\mu^2)(1-x^2)\sin(\mu x)$.

\begin{figure}[htbp]
  \centering
  \includegraphics[width=5cm, angle=90]{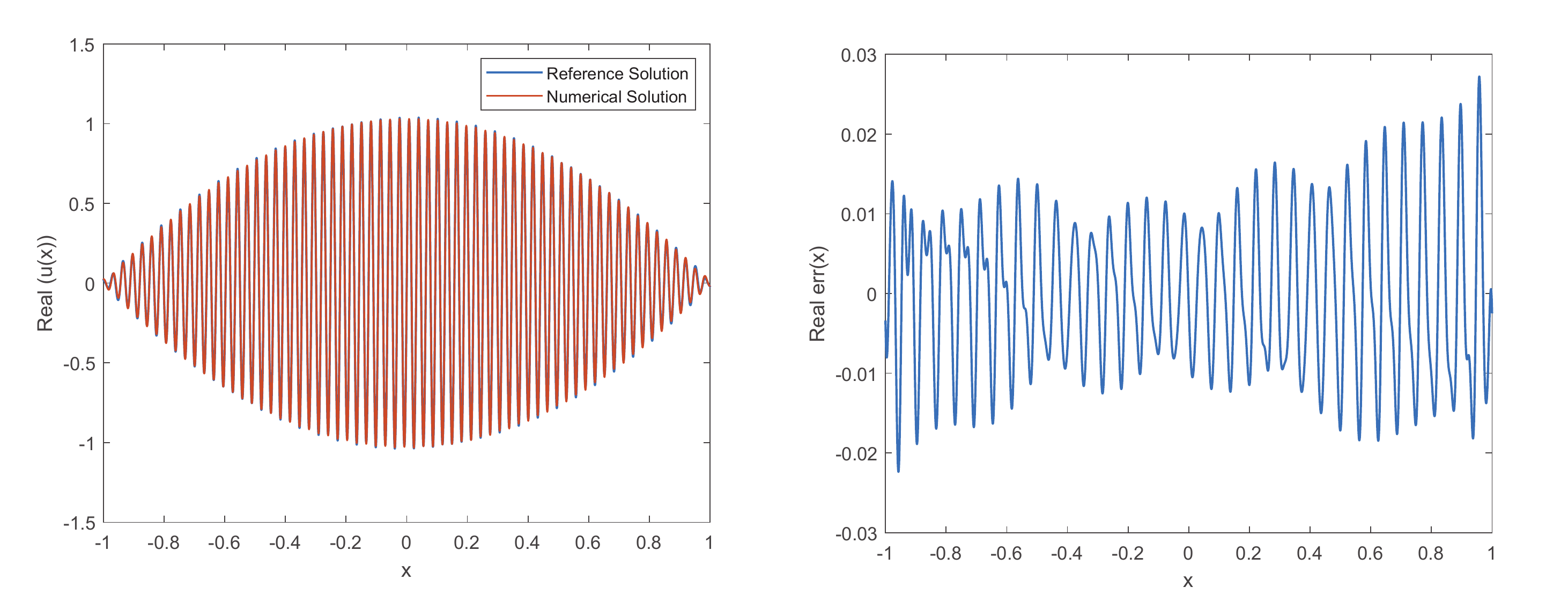}
  \caption{The result of exterior problem using coupled PhaseDNN for the differential equation \eqref{eq:scatter} after 3000 epochs training. Left panel: The real part of numerical and reference solution to exterior problem. Blue line: reference solution. Red line: numerical solution. Right panel: the error of the real part of numerical solution.}\label{Fig:13}
\end{figure}

We first solve the problem with Robin boundary condition \eqref{eq:robin} by the coupled PhaseDNN for the differential equation. The real part of the solution is shown in Fig.\ref{Fig:13}. We set each sub network in \eqref{eq:CPDNNreal} to be a 1-20-20-20-20-1 DNN. Training data set is 3000 evenly spaced points in $[-2,2]$. The training runs 3000 epochs with batchsize 600.

\begin{figure}[htbp]
  \centering
  \includegraphics[width=5cm, angle=90]{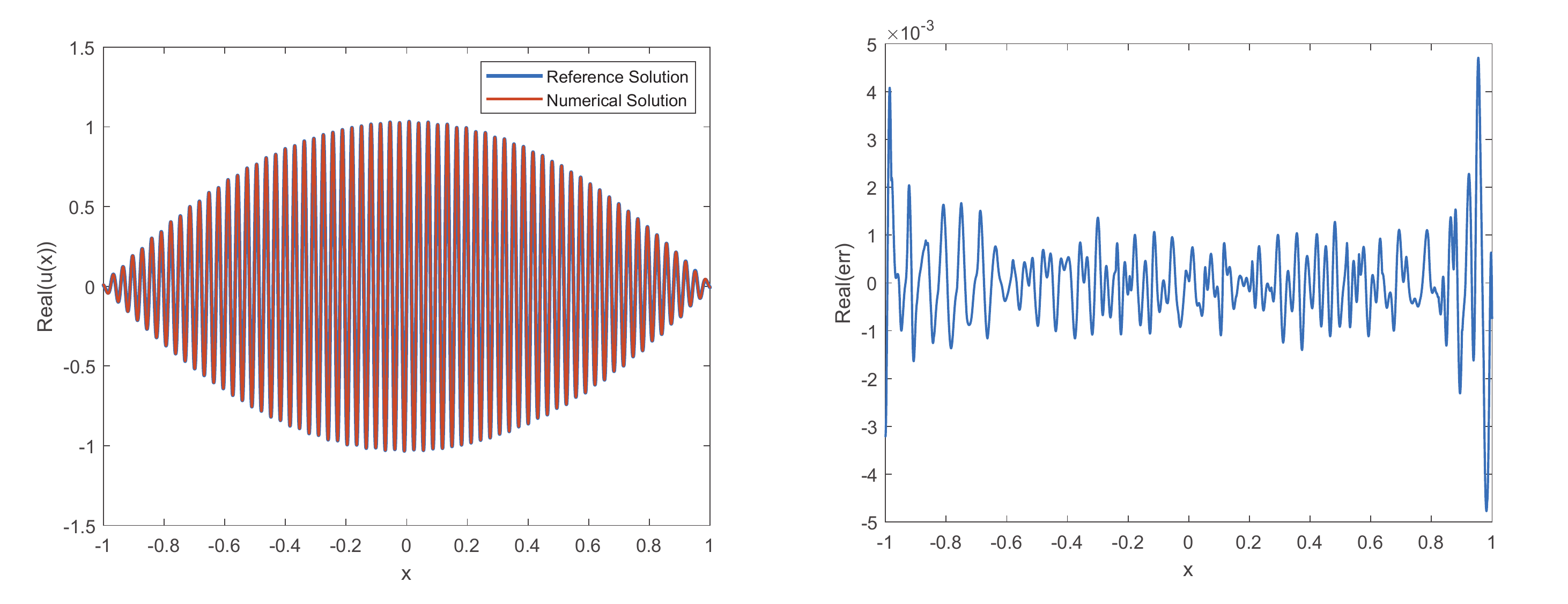}
  \caption{The result of exterior problem using integral equation method after 300 epochs training. Left panel: The real part of numerical and reference solution to exterior problem. Blue line: reference solution. Red line: numerical solution. Right panel: the error of the real part of numerical solution.}\label{Fig:14}
\end{figure}

Again, this problem will be solved by the coupled PhaseDNN with the integral equation  (\ref{inteEq}) (\ref{IntRes}), and the result is given in Fig.\ref{Fig:14}.
The training parameters are set similarly as for the differential equation method. Training runs 300 epochs. Better performance of the coupled PhaseDNN for the integral equation approach (\ref{inteEq}) (\ref{IntRes}), requiring much fewer training epochs,  is shown, compared with the differential equation coupled PhaseDNN.

\section{Conclusion}

In this paper, we have proposed a phase shift DNN to learn high frequency
information by using frequency shifts to convert the high frequency learning
to low frequency one. As shown by various numerical tests, this approach
increases dramatically the capability of the DNN as a viable tool for approximating high frequency
functions and solutions of high frequency wave differential and integral equations in inhomogeneous media.

The optimization problem with the training of DNNs is complex and not much understood during the search
of parameter spaces of the DNNs for a local or global minima. The specific structure of the proposed PhaseDNN seems to guide this search
more efficiently than the common fully connected DNNs.
Meanwhile, our numerical results also show that the PhaseDNN with integral equation formulation of the  high frequency wave problems gives better accuracy to that with
differential equations. These issues will be the subject of future theoretical analysis of
the PhaseDNN. Also for future work, we will further develop the PhaseDNN to handle the high
dimensionality problems from random inhomogeneous media in the wave propagations.

\section*{Appendix}
In this appendix, we will show that under a condition that the weight of the input layer for each sub DNN $T_m(x)$ is small enough, coupled PhaseDNN is approximately equivalent to parallel PhaseDNN in approximating a function during training.

For simplicity, we only consider the 1-D case $d=1$ and assume $|\omega_i-\omega_j| = |i-j|\Delta k$ for all $1\leqslant i,j\leqslant M$ and some $\Delta k>0$.

Generally, the Fourier transform of a target function $f(x)$ and DNN function $T(x)$ may not exists. However, as we are only interested in the target function in a compact domain $\Omega\subset\mathbb{R}^d$. To avoid this technical problem in analysis, we choose a smooth mollifier function $\kappa(x)$ satisfies $\kappa(x)=1$ for $x\in\Omega$ and $\kappa(x)=0$ for $x\in \mathbb{R}^d \backslash \Omega'$, where $\Omega\subset\Omega'\subset\mathbb{R}^d$. We assume $f(x)\kappa(x)\in H^r$ and $T(x)\kappa(x)\in H^r$ for some $r\geq 1$. This assumption ensures the existence of Fourier transform for  $f(x)\kappa(x)$ and $T(x)\kappa(x)$. For simplicity purpose, in the following text, we still use $f(x)$ and $T(x)$ to indicate $f(x)\kappa(x)$ and $T(x)\kappa(x)$.

Although in practice, we usually use the real form of coupled phase DNN \eqref{eq:CPDNNreal}, in analysis, we prefer the complex form \eqref{eq:CPDNNcomplex}. Recall $T_m(x)$ in \eqref{eq:CPDNNcomplex} is complex valued and each DNN $T_m$ in \eqref{eq:CPDNNcomplex} can be written as $T_m(x) = u_m(\eta w_m x+b_m)$, where $\eta>0$ is a small parameter, and $u_m(t)$ are (complex valued) DNN functions except the input layer. Again, in general, $u_m(x)$ is not $\mathbb{L}^2$ function, however, as we are not interest in its behavior at infinity, we can multiply it by a smooth mollifier function $\kappa(x)$ s.t. $\kappa(x)=1$ in a compact domain $\Omega\subset\mathbb{R}$ and $\kappa(x) = 0$ in $\mathbb{R}\backslash\Omega'$. Here, the compact domain $\Omega'$ satisfies $\Omega\subset\Omega'\subset\mathbb{R}$. We assume the new function $u(x)\kappa(x)\in H^r(\mathbb{R})$, $r\geq 1$. For simplicity, in the following text, we still use $u(x)$ stands for the $H^r$ function $u(x)\kappa(x)$. Without losing generality, we can assume $w_m\neq 0$ for all $1\leqslant m\leqslant M$.

We know that $u_m(x,\theta)\in L^2(\mathbb{R})$, where $\theta = \theta(t)$, $t$ stands for training process. It is reasonable to assume that
\begin{asm}
  During the training $t\in [0,T]$, there is a constant $H>0$ s.t
  \[H^{-1}<\inf_{t\in [0,T]}\norm{u_m}_2\leqslant\sup_{t\in [0,T]}\norm{u_m}_2<H\]
  holds for any $1\leqslant m\leqslant M$.
\end{asm}
This assumption is reasonable because if $\norm{u}_2$ has no upper bound means the training blows up while no positive lower bound means this term vanishes. The constant is uniform because $M$ is finite. Furthermore, we can assume
\begin{asm}
  For any $\epsilon>0$, there is constant $A>0$ s.t.
\[\int_{A}^{+\infty}|\hat{u}_m|^2\d k<\epsilon, \mbox{ and } \int_{-\infty}^{-A}|\hat{u}_m|^2\d k<\epsilon\]
holds for any $1\leqslant m\leqslant M$ and $t\in [0,T]$.
\end{asm}

It is straightforward to show that the Fourier transform of \eqref{eq:CPDNNcomplex} yields
\begin{equation}\label{eq:hatT}
  \hat{T}(k) = \sum_{m=1}^{M}\hat{T}_m(k-\omega_m) = \sum_{m=1}^{M}\frac{1}{\eta|w_m|}\exp(-\frac{ib_m(k-\omega_m)}{\eta w_m})\hat{u}_m(\frac{k-\omega_m}{\eta w_m}),
\end{equation}
and
\begin{equation}\label{eq:ltheta}
  L(\theta) = \int_{-\infty}^{\infty}|\hat{f}-\hat{T}|^2\d k = \int_{-\infty}^{\infty}|\hat{f}|^2\d k - 2Re\int_{-\infty}^{\infty}\hat{T}\bar{\hat{f}}\d k + \int_{-\infty}^{\infty}|\hat{T}|^2\d k.
\end{equation}

In eqn. \eqref{eq:ltheta},
\[\int_{-\infty}^{\infty}|\hat{T}|^2\d k = \int_{-\infty}^{\infty}\sum_{m,n=1}^{M}\hat{T}_m(k-\omega_m)\bar{\hat{T}}_n(k-\omega_n)\d k.\]
We will show that under assumption 1 and 2, when $\eta\to 0$,
\begin{equation}\label{eq:crossto0}
  \frac{\int_{-\infty}^{\infty}|T_m(k-\omega_m)||T_n(k-\omega_n)|\d k}{\int_{-\infty}^{\infty}|T_m(k-\omega_m)|^2\d k}\to 0
\end{equation}
for all $m=1,2,\dots M$ and $n\neq m$.

To this end, we first notice that
\begin{equation}\label{eq:bottom}
\begin{aligned}
  \int_{-\infty}^{\infty}|T_m(k-\omega_m)|^2\d k &= \frac{1}{\eta^2 w_m^2}\int_{-\infty}^{\infty}|\hat{u}_m(\frac{k-\omega_m}{\eta w_m})|^2\d k\\
  & = \frac{1}{\eta \abs{w_m}}\int_{-\infty}^{\infty}|\hat{u}_m(t)|^2\d t\\
  & = \frac{\norm{u_m}_2^2}{\eta \abs{w_m}}>\frac{1}{\eta H\abs{w_m}}.
\end{aligned}
\end{equation}

At the same time,

\begin{equation}\label{eq:top}
\begin{aligned}
  \int_{-\infty}^{\infty}|T_m(k-\omega_m)|&|T_n(k-\omega_n)|\d k \\
  &= \frac{1}{\eta^2 \abs{w_m}\abs{w_n}}\int_{-\infty}^{\infty}|\hat{u}_m(\frac{k-\omega_m}{\eta w_m})||\hat{u}_n(\frac{k-\omega_n}{\eta w_n})|\d k\\
  & = \frac{1}{\eta \abs{w_n}}\int_{-\infty}^{\infty}|\hat{u}_m(t)||\hat{u}_n(\frac{w_m}{w_n} t + \frac{\omega_m-\omega_n}{\eta w_n})|\d t\\
  & = \frac{1}{\eta \abs{w_n}} \left( \int_{-\infty}^{-A} + \int_{-A}^{A} + \int_{A}^{\infty}\right).
\end{aligned}
\end{equation}

Applying the Cauchy-Schwardz inequality, we can estimate the first integral by
\begin{equation}\label{eq:Int1}
\begin{aligned}
  \int_{-\infty}^{-A}|\hat{u}_m(t)|&|\hat{u}_n(\frac{w_m}{w_n} t + \frac{\omega_m-\omega_n}{\eta w_n})|\d t \\
  &\leqslant \left(\int_{-\infty}^{-A}|\hat{u}_m(t)|^2\d t\right)^{1/2} \left(\int_{-\infty}^{-A}|\hat{u}_n(\frac{w_m}{w_n} t + \frac{\omega_m-\omega_n}{\eta w_n})|^2\d t\right)^{1/2}\\
  & \leqslant \sqrt{\epsilon\abs{\frac{w_n}{w_m}}}\norm{\hat{u}_n}_2< \epsilon\sqrt{\abs{\frac{w_n}{w_m}}}H.
\end{aligned}
\end{equation}

Similarly, we can also estimate
\begin{equation}\label{eq:Int3}
   \int^{\infty}_{A}|\hat{u}_m(t)||\hat{u}_n(\frac{w_m}{w_n} t + \frac{\omega_m-\omega_n}{\eta w_n})|\d t \leqslant\epsilon \sqrt{\abs{\frac{w_n}{w_m}}}H.
\end{equation}

For the second integral, we have
\begin{equation}\label{eq:Int2}
\begin{aligned}
\int_{-A}^{A}|\hat{u}_m(t)|&|\hat{u}_n(\frac{w_m}{w_n} t + \frac{\omega_m-\omega_n}{\eta w_n})|\d t \\
&\leqslant \left(\int_{-A}^{A}|\hat{u}_m(t)|^2\d t\right)^{1/2}\left(\int_{-A}^{A}|\hat{u}_n(\frac{w_m}{w_n} t + \frac{\omega_m-\omega_n}{\eta w_n})|^2\d t\right)^{1/2} \\
&\leqslant \sqrt{\abs{\frac{w_n}{w_m}}}\norm{u_m}_2\left(\int_{\frac{|\omega_m-\omega_n|}{\eta|w_n|} - \abs{\frac{w_m}{w_n}}A}^{\frac{|\omega_m-\omega_n|}{\eta|w_n|} + \abs{\frac{w_m}{w_n}}A}\abs{\hat{u}_n(t)}^2\d t\right)^{1/2}.
\end{aligned}
\end{equation}
Since $n\neq m$, $\omega_m-\omega_n\neq 0$. When $\eta<\frac{\abs{\omega_m-\omega_n}}{\abs{w_m}+\abs{w_n}}$,
\[\int_{\frac{|\omega_m-\omega_n|}{\eta|w_n|} - \abs{\frac{w_m}{w_n}}A}^{\frac{|\omega_m-\omega_n|}{\eta|w_n|} + \abs{\frac{w_m}{w_n}}A}\abs{\hat{u}_n(t)}^2\d t<\int_{A}^{\infty}\abs{\hat{u}_n(t)}^2\d t<\epsilon.\]
To choose a uniform $\eta$, it is sufficient to set
\[\eta<\min_{m\neq n}\frac{\abs{\omega_m-\omega_n}}{\abs{w_m}+\abs{w_n}}=\frac{\Delta k}{2\max_m\abs{w_m}}>0.\]

Combining eqn \eqref{eq:Int1}, \eqref{eq:Int3}, \eqref{eq:Int2} and \eqref{eq:bottom}, we can conclude that
\[\frac{\int_{-\infty}^{\infty}|T_m(k-\omega_m)||T_n(k-\omega_n)|\d k}{\int_{-\infty}^{\infty}|T_m(k-\omega_m)|^2\d k}\to 0, (\eta\to 0)\]
holds for all $1\leqslant m\neq n \leqslant M$. Thus, when $\eta$ is small enough, we can estimate
\begin{equation}\label{eq:approx1}
\begin{aligned}
\int_{-\infty}^{\infty}|\hat{T}|^2\d k &= \int_{-\infty}^{\infty}\sum_{m,n=1}^{M}\hat{T}_m(k-\omega_m)\bar{\hat{T}}_n(k-\omega_n)\d k \\
&\approx \sum_{m=1}^{M}\int_{-\infty}^{\infty}\abs{\hat{T}_m(k-\omega_m)}^2\d k
\end{aligned}
\end{equation}

We can do decomposition of $\hat{f}(k)=\sum_{m=1}^{M}\hat{f}(k)\chi_m(k)=\sum_{m=1}^{M}\hat{f}_m$, where $\chi_m(k)$ is the indicator function of interval $[\omega_m-\Delta k/2, \omega_m+\Delta k/2]$. It is easy to see $|\hat{f}|^2 = \sum_{m=1}^{M}|\hat{f}_m|^2$ and
\[\int_{-\infty}^{\infty}\hat{T}\bar{\hat{f}}\d k = \sum_{m,n=1}^{M}\int_{-\infty}^{\infty}\hat{T}_m(k-\omega_m)\hat{f}_n(k)\d k.\]
With the same argument in proving eqn \eqref{eq:crossto0}, we can also deduce that if $\hat{f}_n\neq 0$
\begin{equation}\label{eq:crossf}
  \frac{\int_{-\infty}^{\infty}\abs{\hat{T}_m(k-\omega_m)}|\hat{f}_n|\d k}{\int_{-\infty}^{\infty}\abs{\hat{T}_n(k-\omega_n)}|\hat{f}_n|\d k} \to 0
\end{equation}
for all $m=1,2,\dots, M$ and $n\neq m$. Thus
\begin{equation}\label{eq:approx2}
  \int_{-\infty}^{\infty}\hat{T}\bar{\hat{f}}\d k \approx \sum_{m=1}^{M}\int_{-\infty}^{\infty}\hat{T}_m(k-\omega_m)\bar{\hat{f}}_m.
\end{equation}

Substituting approximations \eqref{eq:approx1} and \eqref{eq:approx2} to loss function \eqref{eq:ltheta}, a good approximation of loss function, when $\eta$ is small, is
\begin{equation}\label{eq:apploss}
\begin{aligned}
  L(\theta)&\approx \int_{-\infty}^{\infty}\sum_{m=1}^{M}\abs{\hat{f}_m(k)}^2 - 2Re\hat{T}_m(k-\omega_m)\bar{\hat{f}}_m + \abs{\hat{T}_m(k-\omega_m)}^2\d k\\
  & = \sum_{m=1}^{M}\int_{-\infty}^{\infty}\abs{\hat{f}_m(k) - \hat{T}_m(k-\omega_m)}^2\d k\\
  & = \sum_{m=1}^{M}\int_{-\infty}^{\infty}\abs{\mathcal{F}^{-1}[\hat{f}_m(k+\omega_m)] - T_m(x)}^2\d x\\
  & = \sum_{m=1}^{M}\int_{-\infty}^{\infty}\abs{e^{-i\omega_m x}\mathcal{F}^{-1}[\hat{f}_m](x) - T_m(x)}^2\d x
\end{aligned}
\end{equation}

Here, $e^{-i\omega_m x}\mathcal{F}^{-1}[\hat{f}_m](x)$ is exactly the $f^{\text{shift}}_m(x)$ in \eqref{rjshift}, whose support in frequency space is $[-\Delta k/2, \Delta k/2]$. Hence, the total loss function $L(\theta)$ is approximately the sum of $M$ individual DNNs $T_m$ that learn $f^{\text{shift}}_m(x)$, separately.

%\section*{Acknowledgement}
%
%This work was supported by US Army Research Office (Grant No. W911NF-17-1-0368).

\end{document}